\documentclass{article} 
\usepackage{colm2024_conference}

\usepackage{wrapfig}
\usepackage{anyfontsize}
\usepackage{algorithm}
\usepackage{algpseudocode}
\usepackage{caption}
\usepackage{amssymb}
\usepackage{multirow}
\usepackage{pifont}
\usepackage{microtype}
\usepackage{hyperref}
\usepackage{url}
\usepackage{booktabs}
\usepackage{graphicx}
\definecolor{darkblue}{rgb}{0, 0, 0.5}
\hypersetup{colorlinks=true, citecolor=darkblue, linkcolor=darkblue, urlcolor=darkblue}

\title{Simulating Task-Oriented Dialogues with State Transition Graphs and Large Language Models}

\author{Chris Samarinas$^*$, Pracha Promthaw$^*$, Atharva Nijasure\thanks{Equal contribution} ,\\
\medskip
\textbf{Hansi Zeng, Julian Killingback, Hamed Zamani}\\
Center for Intelligent Information Retrieval\\
University of Massachusetts Amherst\\
United States \\
\texttt{\{csamarinas,ppromthaw,anijasure,hzeng,jkillingback,zamani\}@cs.umass.edu} \\
}

\colmfinalcopy
\begin{document}

\maketitle

\begin{abstract}
This paper explores SynTOD, a new synthetic data generation approach for developing end-to-end Task-Oriented Dialogue (TOD) Systems capable of handling complex tasks such as intent classification, slot filling, conversational question-answering, and retrieval-augmented response generation, without relying on crowdsourcing or real-world data. SynTOD utilizes a state transition graph to define the desired behavior of a TOD system and generates diverse, structured conversations through random walks and response simulation using large language models (LLMs). In our experiments, using graph-guided response simulations leads to significant improvements in intent classification, slot filling and response relevance compared to naive single-prompt simulated conversations. We also investigate the end-to-end TOD effectiveness of different base and instruction-tuned LLMs, with and without the constructed synthetic conversations. Finally, we explore how various LLMs can evaluate responses in a TOD system and how well they are correlated with human judgments. Our findings pave the path towards quick development and evaluation of domain-specific TOD systems. We release our datasets, models, and code for research purposes.
\end{abstract}

\section{Introduction}

Task-Oriented Dialogue (TOD) systems have become increasingly popular in various domains, such as customer support, personal assistants, e-commerce, and enterprise. These systems aim to assist users in accomplishing specific goals through natural language conversations. However, building effective TOD systems requires large amounts of diverse and high-quality training data, which can be expensive and time-consuming to collect \citep{RecentAdvancesAndChallengesInTaskOrientedDialogSystems}. Moreover, existing datasets often lack the complexity and richness needed to develop sophisticated TOD systems that can handle a wide range of user intents, perform slot filling, address information seeking requests, and generate contextually relevant and appealing responses \citep{DialogueQualityAndEmotionAnnotationsForCustomerSupportConversations,CISbook}.

Recent advancements in large language models (LLMs) have shown promising results in various NLP tasks. However, LLMs alone are not sufficient for building effective TOD systems, as they require a set of diverse task-specific training data to perform well in specialized domains \citep{AreLargeLanguageModelsAllYouNeedForTaskOrientedDialogue}. We argue that trivial synthetic data generation approaches using LLMs (for example through single prompt engineering) may not provide the diverse training data required for training robust TOD systems.

To address these challenges, we introduce SynTOD, a simple yet effective framework that takes a \emph{state transition graph} for the target TOD system and translates it into a set of synthetically generated task-oriented dialogues using LLMs. The state transition graph defines the desired behavior of the TOD system and guides the generation of diverse, structured conversations. This approach allows for fine-grained control over the conversation structure, ensuring that the generated data covers a wide range of scenarios and edge cases. Based on our experiments, SynTOD leads up to 37\% improvement in intent classification, 100\% in slot filling and 30\% in response relevance compared to naive single-prompt simulated conversations. By incorporating retrieval augmentation, SynTOD enables the development of TOD systems that can handle complex dialogues that involve navigation, search, result filtering, summarization, and question answering. In this work, we aim at answering the following research questions: \textbf{RQ1:} \textit{How does data generation for TOD with a single prompt compare to graph-guided multi-prompt generation?} \textbf{RQ2:} \textit{How do different LLMs trained on synthetic data perform on end-to-end TOD tasks?} \textbf{RQ3:} \textit{How much synthetic training data is required to build an effective LLM-based TOD system?} and \textbf{RQ4:} \textit{How do various LLMs evaluate response relevance in this setting, and are they correlated with human judgments?}

\begin{wrapfigure}{r}{2.8in}
\centering
\includegraphics[width=2.8in]{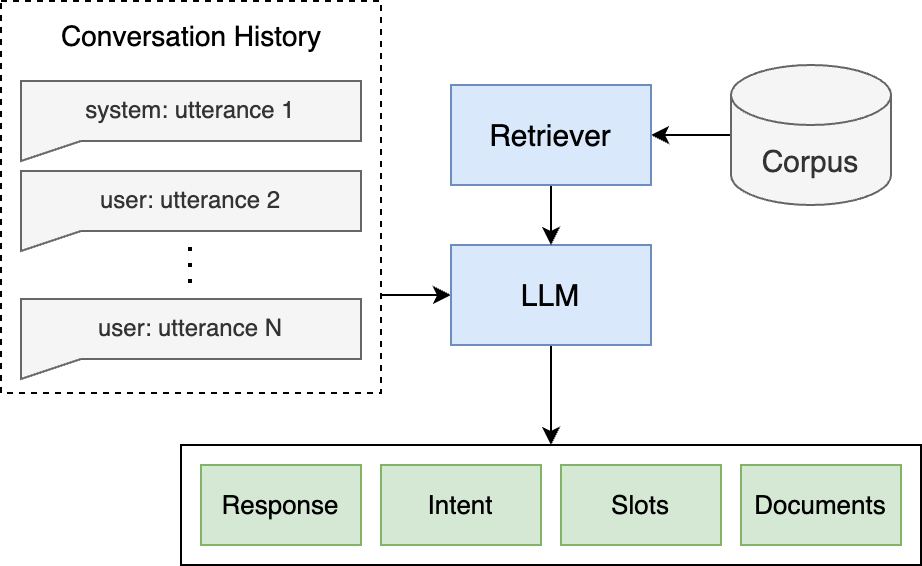}
\caption{Overview of an end-to-end retrieval-augmented TOD system. A LLM and a retriever are the main components. A conversation history is given as input, and response, intent, slots and documents comprise the output system state.}
\label{fig:e2e-system}
\end{wrapfigure} To address these research questions, we conduct extensive experiments in  two new defined domains: cooking and e-commerce assistance. Our Synthetic datasets surpass existing TOD and conversational recommendation datasets in terms of feature richness and diversity, covering 7 different types of interactions compared to other datasets that have up to 4 (see Table \ref{tab:dataset-features}), and allow us to train and evaluate both state tracking and various types of response generation, including attribute-related questions, comparison questions, open-domain questions, result summarization and clarifying questions. We evaluate the performance of various LLM-based TOD systems trained on our synthetic data using both automatic and human evaluations. While there exist other approaches for generating synthetic data for TOD systems, our focus on fully automated graph-guided multi-prompt generation using LLMs with minimal human input sets SynTOD apart. The only required human input is a transition graph and a set of response simulation prompts. Sampling of user intent paths using a state transition graph allows for more structured and diverse conversations. The main contributions of this work are as follows:
\begin{itemize}
    \item We propose SynTOD, a simple framework for generating synthetic training data for end-to-end TOD systems using state transition graphs and LLMs.
    \item We demonstrate the effectiveness of SynTOD in two domains: cooking and e-commerce assistance, generating datasets that surpass existing TOD and conversational recommendation datasets in terms of feature richness and diversity.
    \item We conduct extensive experiments to evaluate the performance of LLM-based end-to-end TOD systems trained on our synthetic data, using both automatic and human evaluation, and provide insights into the four research questions posed above.
    \item We release two synthetic datasets generated using SynTOD, along with a curated test split that can serve as a proxy benchmark for comparing complex TOD systems. We also make our models and implementation publicly available.\footnote{Our data, models and code are available at \url{https://github.com/algoprog/SynTOD}.}
\end{itemize}

By leveraging the power of LLMs and our synthetic data generation framework, SynTOD aims to simplify the development process of sophisticated TOD systems and enhance the user experience in real-world applications. Our approach offers a promising direction for building effective TOD systems in low-resource settings and specialized domains.

\section{Methodology}
In this section, we describe SynTOD, our new framework for synthetic data generation for TOD systems. SynTOD is based on the creation of a state transition graph and the simulation of retrieval-augmented responses using LLMs. 
The primary goal in TOD is to facilitate a user in accomplishing a specific goal through a conversational interface. Effective TOD systems should be capable of understanding the user's intent, filling in the necessary slots, and generating appropriate responses. Given a dialogue history $H = \{u_1, r_1, u_2, r_2, ..., u_t\}$, where $u_i$ and $r_i$ represent the user's and system's utterances respectively at time $i$, a state transition graph $G$, and the parameters of an LLM $M_\theta$, SynTOD should generate a system state $S_{t}$ at time $t$: $S_{t} = \phi(H, G, M_\theta)$. $S_{t}$ is a tuple of three components: the user's intent $I_{t}$ with related attributes (slots) $V_{t}$, and optionally some documents $D_{t}$ that might be retrieved to help generate a response: $S_{t} = (I_{t}, V_t, D_{t})$.

\begin{figure*}[h]
\centering
\includegraphics[width=0.75\textwidth]{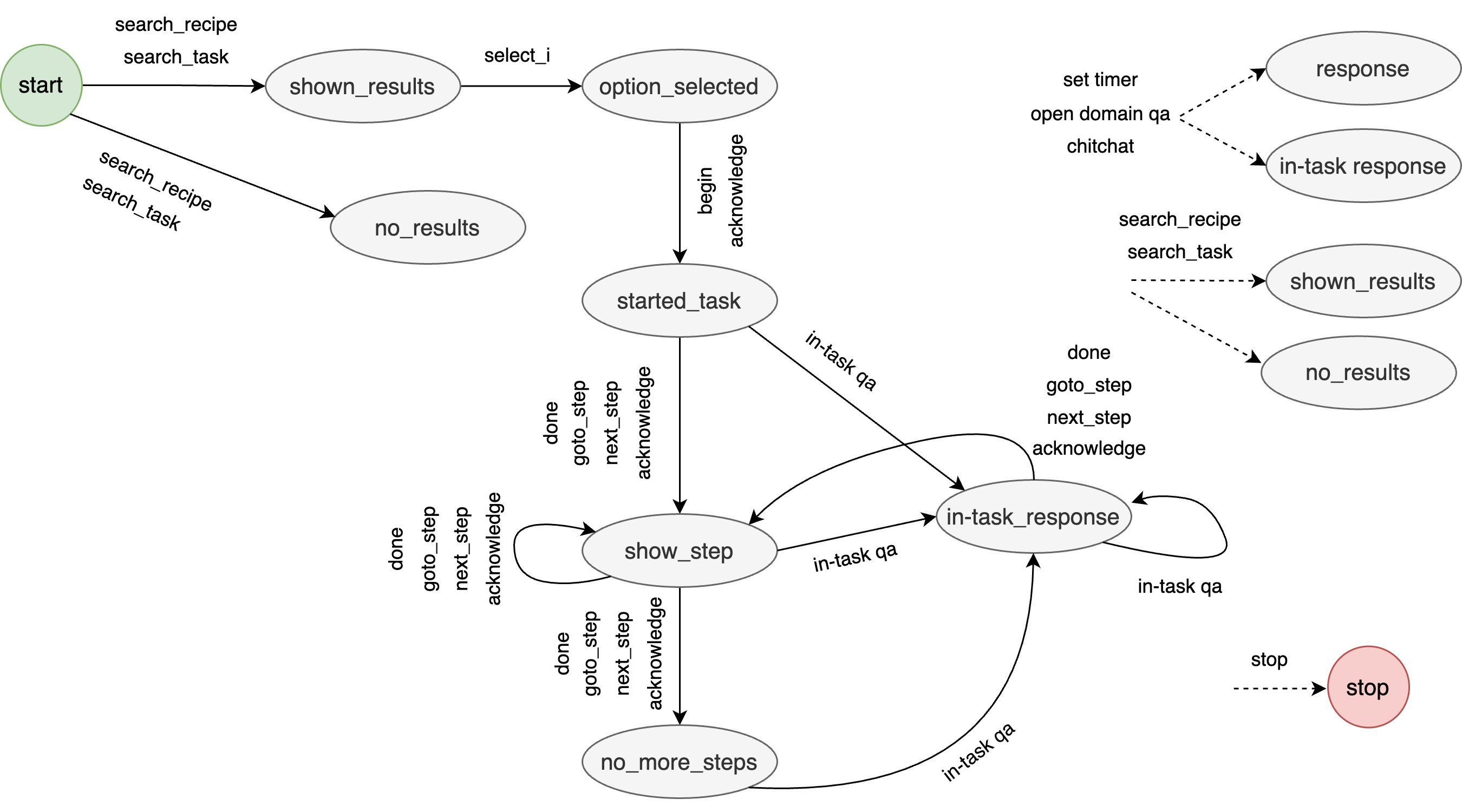}
\caption{The state transition graph we defined for the recipe assistant domain. On the right we see transitions to nodes that are possible from any other state.}
\label{fig:graph}
\end{figure*}
\begin{figure*}[h]
\centering
\includegraphics[width=0.9\textwidth]{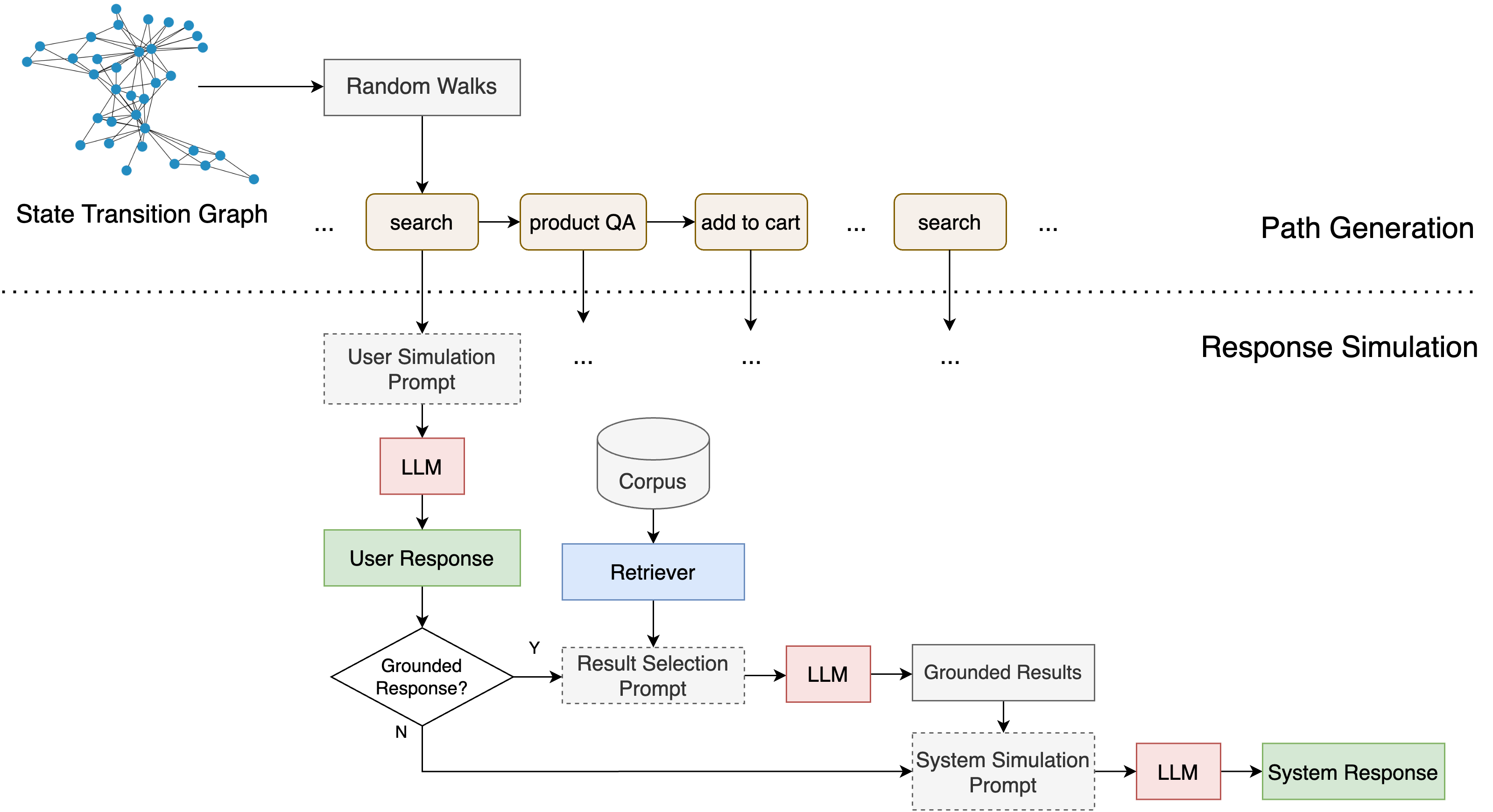}
\caption{Overview of the SynTOD conversation simulation framework.}
\label{fig:syntod}
\end{figure*}

\textbf{State Transition Graph} The first step in our framework involves defining a state transition graph $G$. This graph encapsulates the desired behavior of the TOD system. The nodes in this graph represent the various states of the system, while the edges symbolize the user's intents. The edges also have weights based on the transition probabilities between states. The probabilities can be defined empirically or based on some analysis of existing real conversations in a given domain. The graph is designed to be comprehensive, covering all possible states and transitions that the system may encounter during a conversation and serves as the backbone of the SynTOD framework, guiding the generation of synthetic conversations. An example of a state transition graph for the e-commerce assistance domain can be seen in Figure~\ref{fig:graph}.

\textbf{Data Generation} The data generation process of SynTOD is summarized in Figure~\ref{fig:syntod} and described in detail in Algorithm~\ref{algorithm}. We generate training examples $T_i$ through a process $f$: $T_i = f(D_s,M,P,G)$, where $D_s$ a collection of seed items/documents with relevant metadata, $M$ a pre-trained LLM, $P$ a set of response simulation prompts, one or more per user and system intent, and a state transition graph $G$. The training example $T_i$ generated through the process $f$ includes both the dialogue history $H$ and the corresponding system state $S_{t}$ for each time step $t$. The algorithm starts with a set of random walks generated based on the graph $G$. Each random walk represents a potential conversation path, with each node and edge corresponding to a system state and user intent, respectively. For each node and edge in these random walks, a large language model with a custom prompt is used to simulate a response. A pre-trained LLM is used to generate responses from either the system or the user, depending on the node or edge it is simulating. This process allows for the creation of diverse and realistic synthetic conversations that adhere to the structure defined by the state transition graph. In addition to simulating responses, the SynTOD framework also incorporates retrieval-augmentation. For example, when a user searches for items, a retriever fetches some items from a document corpus, and the LLM selects a subset of them and summarizes them for the user. To simulate the retrieval augmented responses, we use a corpus of items as seed, and then each seed item is treated as the ground truth target for selection and simulation of future responses. This grounded response generation ensures that the responses are contextually accurate and have limited hallucinations, further enhancing the realism and diversity of the synthetic conversations. Moreover, the algorithm also simulates various state variables, such as step numbers, lists, and queries. These state variables are randomly generated to add more variety to the conversations. In particular, queries are generated using special prompts that guide the LLM to create realistic search queries based on the target item metadata.

\textbf{LLM Adaptation for end-to-end TOD} Based on all the training examples $T_i\in T$, we learn the parameters $\theta$ of another LLM, denoted as $M_{\theta}$, for generating the next system state in other real conversational contexts. For training we can employ parameter-efficient fine-tuning methods (PEFT) such as QLoRA \citep{dettmers2023qlora}. Figure~\ref{fig:e2e-system} shows the simplified architecture of an end-to-end TOD system powered by a single fine-tuned LLM ($M_{\theta}$).
\begin{table*}[t]
\centering
\resizebox{\textwidth}{!}{%
\begin{tabular}{lrrrr}
\toprule
 & \multicolumn{2}{c}{\textbf{Cooking Assistance}} & \multicolumn{2}{c}{\textbf{E-commerce Assistance}} \\
\cmidrule(lr){2-3} \cmidrule(lr){4-5}
 & \textbf{Train} & \textbf{Test} & \textbf{Train} & \textbf{Test} \\
\midrule
Total conversations & 2000 & 300 & 2000 & 304 \\
\# utterances per conversation & 35.89 ($\pm$ 10.34) & 29.08 ($\pm$ 10.84) & 23.19 ($\pm$ 7.80) & 22.30 ($\pm$ 7.87) \\
\# tokens per conversation & 1,416 ($\pm$ 386) & 1,448 ($\pm$ 594) & 2,220 ($\pm$ 934) & 1,972 ($\pm$ 815) \\
\# tokens per user utterance & 12.45 ($\pm$ 7.18) & 13.26 ($\pm$ 9.06) & 22.67 ($\pm$ 14.45) & 24.76 ($\pm$ 23.26) \\
\# tokens per system utterance & 57.39 ($\pm$ 26.65) & 74.98 ($\pm$ 36.28) & 89.57 ($\pm$ 54.78)  & 81.73 ($\pm$ 53.86) \\
\bottomrule
\end{tabular}
}
\caption{\label{tab:dataset} Statistics of the graph-guided  dataset for cooking and e-commerce domains.}
\end{table*}
\begin{algorithm*}[ht!]
\small
\caption{SynTOD Conversation Simulation Framework}
\begin{algorithmic}
\Require State transition graph $G$, seed items $D_s$, pre-trained LLM $M$, response simulation prompts $P$, search intents $I_s$, retrieval model $R$, query generation prompt $P_q$
\State Define the state transition graph $G$
\State Generate a set of random walks $W$ based on $G$
\State Initialize an empty set of training examples $T$
\For{each random walk $w \in W$}
    \State Sample a target document $d_t \in D_s$
    \State Initialize dialogue history list $H$, list of system states $S$ and monitored state variables $V$
    \For{each node and edge $(n,e)\in w$}
        \If{previous edge $e_{prev}\in I_s$}
            \State Retrieve $k$ relevant documents $D_r$ for $V(query)$ using retrieval model $R$
            \State $D \leftarrow D_r \cup \{d_t\}$
        \Else
            \State $D \leftarrow \emptyset$
        \EndIf
        \State Generate system response $r_t$ using $M$ and $V,D,d_t$ with prompt $P_n$
        \State Update dialogue history $H \leftarrow H \cup \{r_t\}$
        \If{$e\in I_s$}
            \State Generate search query $q$ using $M$ with prompt $P_q$ and target item $d_t$
            \State Update monitored state variables $V(query) = q$
        \EndIf
        \State Generate user utterance $u_t$ using $M$ and $V$ with prompt $P_e$
        \State Update dialogue history $H \leftarrow H \cup \{u_t\}$
        \State Update system state $S_t = (I_e, V, D)$
        \State Update list of states $S \leftarrow S \cup \{S_t\}$
        
        \State Update other monitored state variables in $V$ (e.g. step counter, lists)
    \EndFor
    \State Create training example $T_i = (H, S)$
    \State $T \leftarrow T \cup \{T_i\}$
\EndFor
\State \textbf{return} $T$
\end{algorithmic}
\label{algorithm}
\end{algorithm*}

\section{Experimental Setup}

\subsection{Datasets}
For our experiments we generated datasets for two domains; cooking and e-commerce assistance. For each domain we have two splits, for training and testing. For the training split generation, we used GPT-4, while for the test split we used LLaMA 2 70B, Mistral Medium and Gemini Pro. For each domain, we generated 2000 conversations for training and 300 (100 from each model) for testing. The statistics of our synthetic dataset can be seen in Table~\ref{tab:dataset}. We can see that in both domains, the conversations tend to be long both in terms of conversation turns and total words, in contrast to existing conversational datasets that tend to be short in both aspects. While we used mostly commercial closed-source models to build our training and testing data with a small cost (around \$300 for recipe and \$500 for e-commerce), it might be possible to generate data of similar quality with recent open-source models such as Mixtral.

\textbf{Cooking Assistance} For the cooking assistance domain, we defined a system that helps the user find relevant recipes from a corpus, answers questions about them and is able to go through the instructions step by step. The system supports two types of queries, generic, when asking for general recipe recommendations based on occasion or preferences (e.g., \textit{I want to make a dessert for Christmas}) and specific, when searching for some specific recipe (e.g., \textit{I want to make fluffy pancakes}). We used 4000 recipes from the Tasty dataset as our corpus \citep{sener2019zero}. The supported intents can be seen in Table~\ref{tab:recipe-domain-intents} and an example conversation in Table~\ref{tab:recipe-dataset-sample-conversation} in the appendix.

\textbf{E-commerce Assistance} For the e-commerce assistance domain, we defined a system that helps users discover products from a corpus, compare, answer questions about them, add or remove them from the cart and finally buy the items in the cart. The system can handle two types of queries, generic and specific. Generic queries do not mention a specific product name (e.g., \textit{I want to buy a smartphone}) and are followed by clarifying question from the system (e.g., \textit{what brand would you like?}). To compile the product corpus, we combined the MAVE dataset \citep{mave} with Amazon reviews \citep{amazon-reviews}, resulting in a dataset comprising 4000 products across 50 diverse categories such as \textit{Shoes} and \textit{Headphones \& Headsets}. This dataset lacked critical information such as delivery locations. To address this, we introduced random location allocations for each product. Shopping often involves multiple simultaneous purchases or comparisons. Consequently, for the conversation simulations we included a primary product seed along with additional randomly selected products for comparison and cart additions. For cart additions we randomly sampled products across different categories, and for comparison lists we kept the same category with the seed product. The definition of the supported intents in this domain can be seen in Table~\ref{tab:ecommerce-dataset-intents} and an example conversation in Table~\ref{tab:ecommerce-dataset-sample-conversation} in the appendix.


\subsection{Training and Inference}

\textbf{Data Format} For training an LLM for end-to-end TOD with our synthetic data, we convert the training examples to text format (similar to ChatML). In the conversations we have the following roles; system, user, suggestions, retriever and item information. The suggestions contain list of generated queries for LLM-assisted retrieval for generic queries (those queries are used to perform retrieval and append results in the retriever role), the retriever role contains result metadata such as item title, description and rating, and the item information role contains all the relevant metadata for a selected recipe or product. Each system and user utterance is separated in two segments. The first segment contains the text response while the second contains system state information in json format, which encompasses the current intent and related slots.

\textbf{Fine-tuning} We fine-tune the language models using the QLoRA \citep{dettmers2023qlora} parameter-efficient method until convergence. The training is performed with a batch size of 2 and 8 gradient accumulation steps, resulting in a total effective batch size of 16. There are 1000 steps in total, and we evaluate the model every 200 steps.  For the LoRA weight, we start with $r=64$ and $\alpha=16$ as the default setting. In the recipe domain, we conduct a hyperparameter search for the LoRA parameters, exploring the following variations: ($r=128$, $\alpha=16$), ($r=64$, $\alpha=32$), and ($r=64$, $\alpha=8$). The best-performing settings for the e-commerce domain are $\alpha=32$ and $r=64$ at the step 600 (70.59\% of the training data), while for the recipe domain, $\alpha=8$ and $r=64$ at the steps 100 (11.76\% of the training data), yield the best results.

\textbf{Nearest-neighbor Intent Selection} The intent prediction is performed by the LLM fine-tuned on our synthetic data. However, the generated intent text itself is not always consistent. For example, the "add\_to\_cart" intent could be generated as "ADD TO CART" or "select\_i\_add\_to\_cart". To address this issue and improve the stability of intent generation, we incorporate a dot product nearest neighbor approach on the generated intent,\footnote{We used \href{https://huggingface.co/sentence-transformers/all-MiniLM-L6-v2}{all-MiniLM-L6-v2} as our embedding model} mapping it to the closest available intent in our predefined set. This ensures that the model produces the expected intent it intended to generate.

\section{Results and Analyses}

\begin{figure}[h]
    \centering
    \begin{minipage}{0.37\textwidth}
        \centering
        \small
        \begin{tabular}{lcc}
            \toprule
            \textbf{Dataset} & \textbf{Graph} & \textbf{Self-BLEU} \\ 
            \midrule
            \multirow{2}{*}{Cooking} & \ding{51} & \textbf{0.7850} \\
             & \ding{55} & 0.8101 \\ 
            \midrule
            \multirow{2}{*}{Ecommerce} & \ding{51} & \textbf{0.8123} \\
             & \ding{55} & 0.8932 \\ 
            \bottomrule
        \end{tabular}
        \captionof{table}{Diversity of data generated with and without a graph.}
        \label{tab:diversity}
    \end{minipage}\hfill
    \begin{minipage}{0.53\textwidth}
        \centering
        \includegraphics[width=\textwidth]{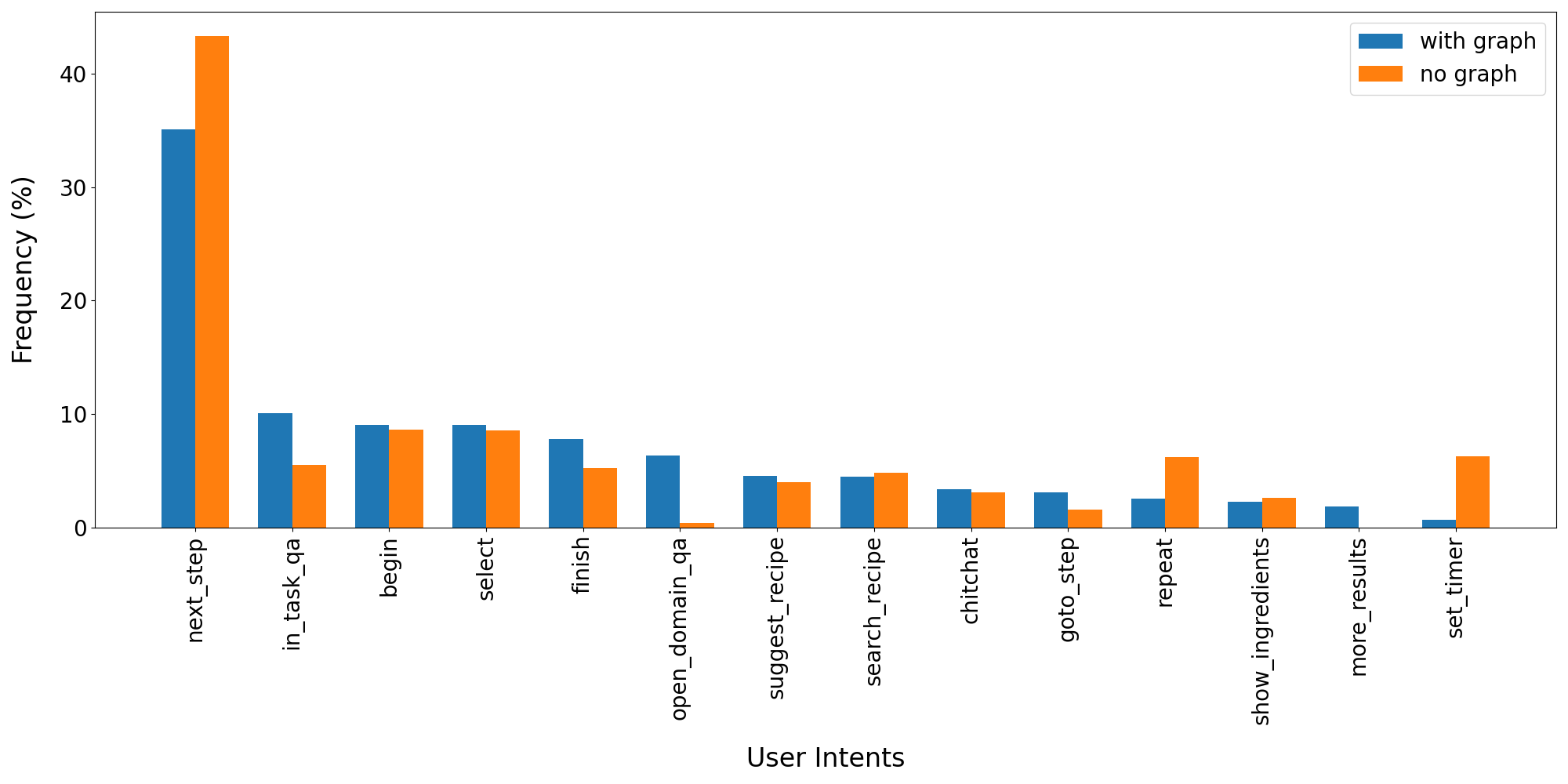}
        \captionof{figure}{Frequency distribution of user intents for the recipe domain with and without state transition graph.}
        \label{fig:intents-comp}
    \end{minipage}
\end{figure}

\textbf{RQ1: How does data generation with a single prompt compare to graph-guided multi-prompt generation?} To address the first research question, we compared the diversity and coverage of the generated data using a single prompt versus graph-guided multi-prompt generation. The results demonstrate that graph-guided multi-prompt generation leads to more diverse data and better coverage of intents. Figure~\ref{fig:intents-comp} illustrates that without the graph, some intents have near-zero frequency while others have higher frequency than desired. Additionally, using a single prompt results in less diverse utterances, as evidenced by the higher Self-BLEU \citep{SELFBLEU} scores (see Table~\ref{tab:diversity}). These findings highlight the importance of the state transition graph in enforcing the desired distribution of user intents and promoting utterance diversity. The model trained with data generated from the transition graph significantly outperforms the model trained without the graph in all intent classification and slot filling metrics, including micro precision, recall, and F1 score (see Table~\ref{tab:performance_comparison}). It is worth noting that for slot filling evaluation, the metrics might be misleading to an extent, because in some cases the predicted slots are correct but with small variations (e.g. 'pancake recipe' instead of 'pancake'). Figures~\ref{fig:recipe_domain_matrix} and \ref{fig:ecommerce_domain_matrix} in the appendix, further illustrate the improvement in accuracy across various intents when using graph-guided training data. These results demonstrate the effectiveness of the state transition graph in generating higher quality training data for intent classification and slot filling. Response relevance evaluation using human annotators, Mixtral and OpenChat 3.5 provides insights into the quality of the generated responses (see Table~\ref{tab:performance_comparison_response_llm}). We defined 3 relevance levels; 0 means the generated response is irrelevant or has major issues, 1 is relevant but worse than the ground truth and 2 means as good as the ground truth. For the human annotation, we asked 3 participants to give one of these labels for each response given the conversation history. We used majority voting and chose the lowest rating in case of ties. The results show that the model trained with graph-guided data achieves higher average relevance scores compared to the model trained without the graph, indicating the superiority of the graph-guided approach in generating high-quality responses.

\begin{table}[htbp]
\centering
\begin{tabular}{lcccc}
\toprule
\multirow{2}{*}{\textbf{Metric}} & \multicolumn{2}{c}{\textbf{Cooking}} & \multicolumn{2}{c}{\textbf{E-commerce}} \\
\cmidrule(lr){2-3} \cmidrule(lr){4-5}
 & no graph & with graph & no graph & with graph \\
\midrule
intent accuracy & 0.7620 & \textbf{0.9580}* & 0.6719 & \textbf{0.8296}* \\
intent precision & 0.7087 & \textbf{0.9344}* & 0.6534 & \textbf{0.8030}* \\
intent recall & 0.6591 & \textbf{0.9477}* & 0.6585 & \textbf{0.8435}* \\
intent f1 & 0.6830 & \textbf{0.9410}* & 0.6560 & \textbf{0.8228}* \\
\midrule
slot precision & 0.5682 & \textbf{0.7417}* & 0.3157 & \textbf{0.5693}* \\
slot recall & 0.6001 & \textbf{0.7343}* & 0.2789 & \textbf{0.5882}* \\
slot f1 & 0.5561 & \textbf{0.7180}* & 0.2840 & \textbf{0.5690}* \\
\bottomrule
\end{tabular}
\caption{Performance comparison of models for intent classification and slot filling with and without a state transition graph. Statistical significance is denoted by $^{*}$ ($p < 0.01$).}
\label{tab:performance_comparison}
\end{table}

\textbf{RQ2: How do different LLMs perform on end-to-end TOD?} To investigate the second research question, we tested three different LLMs with similar architecture and number of parameters (7B) on end-to-end TOD tasks. The results, as shown in Table~\ref{tab:performance_comparison_llms}, indicate that there is no significant difference in performance across different models (Llama 2 vs. Mistral). Furthermore, instruction-tuning, also known as alignment, does not seem to have a significant effect on performance.

\textbf{RQ3: How much synthetic training data is required?} To address the third research question, we used the 2K generated examples (We saved 300 examples for validation and 1700 examples for training) for each domain and observed the model's convergence using different proportions of the generated data. Our results showed that the model converges using just 11.76\% of the examples for the recipe domain (200 examples) and 70.59\% (1200 examples) for the e-commerce domain.

\textbf{RQ4: How do various LLMs evaluate response relevance, and are they correlated with human judgments?} To investigate the fourth research question, we tested two LLMs for automated response relevance evaluation (Mixtral and OpenChat 3.5) and compared their assessments with human judgments. We calculated the Cohen's $\kappa$ for each pair and we observed 17.14\% for Mixtral and 10.34\% for OpenChat in cooking domain and 18.29\% for Mixtral and 14.9\% for OpenChat in e-commerce domain. While Mixtral has much higher correlation than OpenChat with human judgments, it is still not high enough to be considered reliable for automatic evaluation. However, we also calculated the agreement of GPT-4 annotations with human judgments and observed 64.29\% for cooking domain and 43.31\% for e-commerce. It is obvious that correlation of judgments increases with the number of LLM parameters, and GPT-4 or other models of that scale might be viable for automatic evaluation, however not easily reproducible.

\begin{table}[htbp!]
\centering
\begin{tabular}{@{}llccccl@{}}
\toprule
\multirow{2}{*}{\textbf{Domain}} & \multirow{2}{*}{\textbf{Evaluator}} & \multirow{2}{*}{\textbf{Graph}} & \multicolumn{3}{c}{\textbf{Relevance (\%)}} & \multirow{2}{*}{\textbf{Avg. Relevance}} \\
\cmidrule(lr){4-6}
 &  &  & 0 & 1 & 2 &  \\
\midrule
\multirow{6}{*}{Cooking} & Human & \ding{55} & 7.86 & 7.79 & 84.35 & 1.7649\\
 & Human & \ding{51} & 3.56 & 7.12 & 89.23 & \textbf{1.8558}* (+5.15\%)\\
 \cmidrule(lr){2-7}
 & Mixtral 8x7B Instruct & \ding{55} & 0.62 & 13.25 & 86.14 & 1.8553\\
 & Mixtral 8x7B Instruct & \ding{51} & 0.34 & 3.43 & 96.22 & \textbf{1.9587}$^{*}$ (+5.57\%)\\
 \cmidrule(lr){2-7}
 & OpenChat 3.5 & \ding{55} & 1.06 & 7.72 & 91.22 & 1.9016\\
 & OpenChat 3.5 & \ding{51} & 0.62 & 1.96 & 97.43 & \textbf{1.9682}$^{*}$ (+3.50\%)\\
\midrule
\multirow{6}{*}{E-commerce} & Human & \ding{55} & 18.81 & 14.46 & 67.73 & 1.4792 \\
 & Human & \ding{51} & 1.32 & 5.01 & 93.67 & \textbf{1.9235}* (+30.04\%)\\
 \cmidrule(lr){2-7}
 & Mixtral 8x7B Instruct & \ding{55} & 0.95 & 3.70 & 95.35 & 1.9440\\
 & Mixtral 8x7B Instruct & \ding{51} & 0.56 & 1.91 & 97.54 & \textbf{1.9699}$^{*}$ (+1.33\%)\\
 \cmidrule(lr){2-7}
 & OpenChat 3.5 & \ding{55} & 0.67 & 6.57 & 92.76 & 1.9209\\
 & OpenChat 3.5 & \ding{51} & 0.33 & 2.28 & 97.38 & \textbf{1.9704}$^{*}$ (+2.58\%)\\
\bottomrule
\end{tabular}
\caption{Evaluation results of responses from our fine-tuned TOD model based on Llama 2 7B, using synthetic data guided with and without a transition graph for the cooking and e-commerce domains. The superscript $^{*}$ denotes statistical significance compared to the setting without graph in terms of two-tailed paired t-test with $p < 0.01$.}
\label{tab:performance_comparison_response_llm}
\end{table}

\begin{table}[h!]
  \resizebox{\textwidth}{!}{%
\centering
\begin{tabular}{@{}llcccccccc@{}}
\toprule
\multirow{2}{*}{\textbf{Domain}} & \multirow{2}{*}{\textbf{Model}} & \multicolumn{3}{c}{\textbf{Intent}} & \multicolumn{3}{c}{\textbf{Slot}} \\
\cmidrule(lr){3-5} \cmidrule(lr){6-8}
 &  & \textbf{Prec.} & \textbf{Rec.} & \textbf{F1} & \textbf{Prec.} & \textbf{Rec.} & \textbf{F1} & \\
\midrule
\multirow{3}{*}{Cooking} & Llama 2 7B & \textbf{0.9499} & 0.9265 & 0.9211 & \textbf{0.7508}* & \textbf{0.7395}* & \textbf{0.7249}* \\
 & Mistral 7B & 0.9498 & 0.9344 & \textbf{0.9248}* & 0.7231 & 0.7154 & 0.7002 \\
 & OpenChat 3.5 & 0.9471 & \textbf{0.9405}* & 0.9213 & 0.7362 & 0.7304 & 0.7134 \\
\midrule
\multirow{3}{*}{E-commerce} & Llama 2 7B & \textbf{0.8212}* & 0.8348 & 0.8184 & 0.5518 & 0.5692 & 0.5503 \\
 & Mistral 7B & 0.8163 & \textbf{0.8405}* & \textbf{0.8243}* & \textbf{0.6046}* & \textbf{0.6072}* & \textbf{0.5984}* \\
 & OpenChat 3.5 & 0.8099 & 0.8278 & 0.8220 & 0.5764 & 0.5814 & 0.5713 \\
\bottomrule
\end{tabular}
}
\caption{Comparison of 3 different fine-tuned LLMs for end-to-end TOD using synthetic data guided with a state transition graph for the cooking and e-commerce domains. The superscript $^{*}$ denotes statistical significance in terms of two-tailed paired t-test with $p < 0.01$.}
\label{tab:performance_comparison_llms}
\end{table}

\begin{table}[h!]
\centering
\begin{tabular}{@{}llcccc@{}}
\toprule
\multirow{2}{*}{\textbf{Domain}} & \multirow{2}{*}{\textbf{Model}} & \multicolumn{3}{c}{\textbf{Relevance (\%)}} & \multirow{2}{*}{\textbf{Avg. Relevance}} \\
\cmidrule(lr){3-5}
 &  & 0 & 1 & 2 &  \\
\midrule
\multirow{3}{*}{Cooking} & Llama 2 7B & 0.34 & 3.43 & 96.22 & 1.9587\\
 & Mistral 7B & 0.19 & 3.41 & 96.40 & \textbf{1.9621}\\
 & OpenChat 3.5 & 0.16 & 3.66 & 96.18 & 1.9602\\
\midrule
\multirow{3}{*}{E-commerce} & Llama 2 7B & 0.56 & 1.91 & 97.54 & \textbf{1.9699}\\
 & Mistral 7B & 0.77 & 2.23 & 97.00 & 1.9623\\
 & OpenChat 3.5 & 0.95 & 2.39 & 96.66 & 1.9561\\
\bottomrule
\end{tabular}
\caption{Evaluation results of text responses from 3 different fine-tuned LLMs for end-to-end TOD in the cooking and e-commerce domains using synthetic data guided with a state transition graph and Mixtral as the evaluator.}
\label{tab:performance_comparison_response}
\end{table}


\section{Related Work}

TOD data can be collected and created through various approaches. The most straightforward method is to have users interact with an actual system \citep{TheDialogStateTrackingChallenge, LetsGoPublic, IncrementalOnlineAdaptationOfPOMDPbased}. However, this approach is limited by the capabilities of the existing system. The Wizard-of-Oz (WOZ) approach \citep{WizardOfOz} addresses this issue by having humans play the role of the system \citep{ANetworkBasedEndToEndTrainableTaskOrientedDialogueSystem, FramesACorpusForAddingMemoryToGoalOrientedDialogueSystems, MultiWozALargeScaleMultiDomainWIzardOfOZDatasetForTaskOrientedDialogueModelling, TaskOrientedDialogueAsDataflowSynthesis, Taskmaster1TowardARealisticAndDiverseDialogDataset}, allowing for more complex simulated behavior at the cost of increased human effort and less control over system responses. 
To address the limitations of human-driven data collection, some works have employed automated approaches to simulate user-system interactions. These methods generate conversation outlines based on finite state machines \citep{BootstrappingANeuralConversationalAgentWithDialogueSelfPlayCrowdsourcingAndOnLineReinforcementLearning}, probabilistic automata \citep{TowardsScalableMultiDomainConversationalAgentsTheSchemaGuidedDialogueDataset}, or heuristic rules \citep{DialogSimulationWithRealisticVariationsForTrainingGoalOrientedConversationalSystems}. The outlines are then converted into rough natural text using templates, which are further refined by crowd-source workers to enhance diversity and naturalness. A related approach \citep{AlexaConversationsAnExtensibleDataDrivenApproach} uses seed dialogues that are expanded with a simulator to create dialogues, with the option of using crowd-source workers to add response diversity. Although more cost-effective than fully human-generated dialogues, these approaches still require human intervention and may suffer from unrealistic conversational flows.

Recently, efforts have been made to minimize human effort by leveraging generative models. \citet{SimulatedChatsForBuildingDialogSystemsLearningToGenerateConversationsFromInstructions} used crowd-worker data to train a GPT-2 \citep{GPT2} model to simulate both user and system roles. LLMs have also been employed via prompting and fine-tuning to produce TOD systems \citep{ControllableDialogueSimulationWithInContextLearning, SGPTODBuildingTaskBotsEffortlesslyViaSchemaGuidedLlmPrompting, BootstrappingLlmBasedTaskOrientedDialogueAgentsViaSelfTalk}.
Previous datasets primarily focus on canonical TOD scenarios where users seek assistance with specific needs or tasks. To better reflect real-world conversations, recent efforts have integrated other conversation types, such as open-domain conversation, into TOD datasets \citep{FusingTaskOrientedAndOpenDomainDialoguesInConversationalAgents}. Some datasets also incorporate more challenging tasks or contexts, such as search and recommendation \citep{Taskmaster2} and document-grounded conversations \citep{Multidoc2DialModelingDialoguesGroundedInMultipleDocuments}.
Traditionally, TOD systems have relied on multiple components for intent classification, slot filling, state tracking, and handcrafted rules \citep{EndToEndTaskOrientedSurvey}. To simplify this complex architecture, end-to-end systems have been developed where all necessary functions are trained simultaneously. Early efforts used memory networks \citep{LearningEndToEndGoalOrientedDialog} and sequence-to-sequence \citep{ANetworkBasedEndToEndTrainableTaskOrientedDialogueSystem} approaches. More recently, neural language models and LLMs have been employed \citep{UbarTowardsFullyEndToEndTaskOrientedDialogSystemWithGpt2, AreLargeLanguageModelsAllYouNeedForTaskOrientedDialogue, DiaggptAnLlmBasedChatbotWithAutomaticTopicManagementForTaskOrientedDialogue, InstructtodsLargeLanguageModelsForEndToEndTaskOrientedDialogueSystems}.

\section{Conclusion \& Future Work}
In this paper, we introduced SynTOD, a novel framework for generating synthetic data for end-to-end retrieval-augmented task-oriented dialogue systems. Our approach leverages state transition graphs to define the desired behavior of the system and LLMs to simulate diverse and realistic conversations. We demonstrated the effectiveness of SynTOD in two domains: recipe assistance and e-commerce assistance.

The synthetic data generated using our framework exhibited superior diversity compared to data generated from a single prompt. Furthermore, the models trained on the graph-guided synthetic data significantly outperformed those trained on data generated without the graph in both intent classification and slot filling tasks as well as response relevance. We tested three different LLMs on end-to-end TOD with our synthetic data, and we didn't observe significant differences for models of the same scale (7B parameters), and instruction tuning does not seem beneficial. In terms of convergence, we observed that 200-1000 examples might be sufficient depending on the complexity of the chosen domain. For automatic response evaluation, LLMs with larger number of parameters have higher correlation with human judgements, and based on our experiments, only GPT-4 (or potentially models of similar size) could be considered as reliable as human annotators in zero-shot setting. 

Our work showcases the potential of synthetic data generation for building complex, retrieval-augmented task-oriented dialogue systems without relying on expensive real-world data collection. The SynTOD framework simplifies the development process and enhances the user experience by enabling the creation of sophisticated systems capable of handling a wide range of features and tasks. To facilitate further research and development in this area, we release the synthetic datasets used in our experiments, along with a curated test split that can serve as a proxy benchmark for evaluating retrieval-augmented task-oriented dialogue systems. We believe that our contributions will pave the way for more advanced and user-centric conversational AI systems in the future.


\section*{Acknowledgements}
This work was supported in part by the Center for Intelligent Information Retrieval, in part by the Amazon Alexa Prize Competition, in part by Adobe, in part by NSF grant \#2143434, and in part by the Office of Naval Research contract \#N000142212688. Any opinions, findings and conclusions or recommendations expressed in this material are those of the authors and do not necessarily reflect those of the sponsor.

\bibliography{colm2024_conference}

\begin{thebibliography}{46}
\providecommand{\natexlab}[1]{#1}
\providecommand{\url}[1]{\texttt{#1}}
\expandafter\ifx\csname urlstyle\endcsname\relax
  \providecommand{\doi}[1]{doi: #1}\else
  \providecommand{\doi}{doi: \begingroup \urlstyle{rm}\Url}\fi

\bibitem[Acharya et~al.(2021)Acharya, Adhikari, Agarwal, Auvray, Belgamwar, Biswas, Chandra, Chung, Fazel{-}Zarandi, Gabriel, Gao, Goel, Hakkani{-}T{\"{u}}r, Jezabek, Jha, Kao, Krishnan, Ku, Goyal, Lin, Liu, Mandal, Metallinou, Naik, Pan, Paul, Perera, Sethi, Shen, Strom, and Wang]{AlexaConversationsAnExtensibleDataDrivenApproach}
Anish Acharya, Suranjit Adhikari, Sanchit Agarwal, Vincent Auvray, Nehal Belgamwar, Arijit Biswas, Shubhra Chandra, Tagyoung Chung, Maryam Fazel{-}Zarandi, Raefer Gabriel, Shuyang Gao, Rahul Goel, Dilek Hakkani{-}T{\"{u}}r, Jan Jezabek, Abhay Jha, Jiun{-}Yu Kao, Prakash Krishnan, Peter Ku, Anuj Goyal, Chien{-}Wei Lin, Qing Liu, Arindam Mandal, Angeliki Metallinou, Vishal~Ishwar Naik, Yi~Pan, Shachi Paul, Vittorio Perera, Abhishek Sethi, Minmin Shen, Nikko Strom, and Eddie Wang.
\newblock Alexa conversations: An extensible data-driven approach for building task-oriented dialogue systems.
\newblock In Avi Sil and Xi~Victoria Lin (eds.), \emph{Proceedings of the 2021 Conference of the North American Chapter of the Association for Computational Linguistics: Human Language Technologies: Demonstrations, {NAACL-HLT} 2021, Online, June 6-11, 2021}, pp.\  125--132. Association for Computational Linguistics, 2021.
\newblock \doi{10.18653/V1/2021.NAACL-DEMOS.15}.
\newblock URL \url{https://doi.org/10.18653/v1/2021.naacl-demos.15}.

\bibitem[Andreas et~al.(2020)Andreas, Bufe, Burkett, Chen, Clausman, Crawford, Crim, DeLoach, Dorner, Eisner, Fang, Guo, Hall, Hayes, Hill, Ho, Iwaszuk, Jha, Klein, Krishnamurthy, Lanman, Liang, Lin, Lintsbakh, McGovern, Nisnevich, Pauls, Petters, Read, Roth, Roy, Rusak, Short, Slomin, Snyder, Striplin, Su, Tellman, Thomson, Vorobev, Witoszko, Wolfe, Wray, Zhang, and Zotov]{TaskOrientedDialogueAsDataflowSynthesis}
Jacob Andreas, John Bufe, David Burkett, Charles Chen, Josh Clausman, Jean Crawford, Kate Crim, Jordan DeLoach, Leah Dorner, Jason Eisner, Hao Fang, Alan Guo, David Hall, Kristin Hayes, Kellie Hill, Diana Ho, Wendy Iwaszuk, Smriti Jha, Dan Klein, Jayant Krishnamurthy, Theo Lanman, Percy Liang, Christopher~H. Lin, Ilya Lintsbakh, Andy McGovern, Aleksandr Nisnevich, Adam Pauls, Dmitrij Petters, Brent Read, Dan Roth, Subhro Roy, Jesse Rusak, Beth Short, Div Slomin, Ben Snyder, Stephon Striplin, Yu~Su, Zachary Tellman, Sam Thomson, Andrei Vorobev, Izabela Witoszko, Jason~Andrew Wolfe, Abby Wray, Yuchen Zhang, and Alexander Zotov.
\newblock Task-oriented dialogue as dataflow synthesis.
\newblock \emph{Trans. Assoc. Comput. Linguistics}, 8:\penalty0 556--571, 2020.
\newblock \doi{10.1162/TACL\_A\_00333}.
\newblock URL \url{https://doi.org/10.1162/tacl\_a\_00333}.

\bibitem[Asri et~al.(2017)Asri, Schulz, Sharma, Zumer, Harris, Fine, Mehrotra, and Suleman]{FramesACorpusForAddingMemoryToGoalOrientedDialogueSystems}
Layla~El Asri, Hannes Schulz, Shikhar Sharma, Jeremie Zumer, Justin Harris, Emery Fine, Rahul Mehrotra, and Kaheer Suleman.
\newblock Frames: a corpus for adding memory to goal-oriented dialogue systems.
\newblock In Kristiina Jokinen, Manfred Stede, David DeVault, and Annie Louis (eds.), \emph{Proceedings of the 18th Annual SIGdial Meeting on Discourse and Dialogue, Saarbr{\"{u}}cken, Germany, August 15-17, 2017}, pp.\  207--219. Association for Computational Linguistics, 2017.
\newblock \doi{10.18653/V1/W17-5526}.
\newblock URL \url{https://doi.org/10.18653/v1/w17-5526}.

\bibitem[Bordes et~al.(2017)Bordes, Boureau, and Weston]{LearningEndToEndGoalOrientedDialog}
Antoine Bordes, Y{-}Lan Boureau, and Jason Weston.
\newblock Learning end-to-end goal-oriented dialog.
\newblock In \emph{5th International Conference on Learning Representations, {ICLR} 2017, Toulon, France, April 24-26, 2017, Conference Track Proceedings}. OpenReview.net, 2017.
\newblock URL \url{https://openreview.net/forum?id=S1Bb3D5gg}.

\bibitem[Budzianowski et~al.(2018)Budzianowski, Wen, Tseng, Casanueva, Ultes, Ramadan, and Ga{\v{s}}i{\'c}]{MultiWozALargeScaleMultiDomainWIzardOfOZDatasetForTaskOrientedDialogueModelling}
Pawe{\l} Budzianowski, Tsung-Hsien Wen, Bo-Hsiang Tseng, I{\~n}igo Casanueva, Stefan Ultes, Osman Ramadan, and Milica Ga{\v{s}}i{\'c}.
\newblock {M}ulti{WOZ} - a large-scale multi-domain {W}izard-of-{O}z dataset for task-oriented dialogue modelling.
\newblock In Ellen Riloff, David Chiang, Julia Hockenmaier, and Jun{'}ichi Tsujii (eds.), \emph{Proceedings of the 2018 Conference on Empirical Methods in Natural Language Processing}, pp.\  5016--5026, Brussels, Belgium, October-November 2018. Association for Computational Linguistics.
\newblock \doi{10.18653/v1/D18-1547}.
\newblock URL \url{https://aclanthology.org/D18-1547}.

\bibitem[Byrne et~al.(2019)Byrne, Krishnamoorthi, Sankar, Neelakantan, Goodrich, Duckworth, Yavuz, Dubey, Kim, and Cedilnik]{Taskmaster1TowardARealisticAndDiverseDialogDataset}
Bill Byrne, Karthik Krishnamoorthi, Chinnadhurai Sankar, Arvind Neelakantan, Ben Goodrich, Daniel Duckworth, Semih Yavuz, Amit Dubey, Kyu{-}Young Kim, and Andy Cedilnik.
\newblock Taskmaster-1: Toward a realistic and diverse dialog dataset.
\newblock In Kentaro Inui, Jing Jiang, Vincent Ng, and Xiaojun Wan (eds.), \emph{Proceedings of the 2019 Conference on Empirical Methods in Natural Language Processing and the 9th International Joint Conference on Natural Language Processing, {EMNLP-IJCNLP} 2019, Hong Kong, China, November 3-7, 2019}, pp.\  4515--4524. Association for Computational Linguistics, 2019.
\newblock \doi{10.18653/V1/D19-1459}.
\newblock URL \url{https://doi.org/10.18653/v1/D19-1459}.

\bibitem[Byrne et~al.(2020)Byrne, Krishnamoorthi, Ganesh, Dubey, Kim, and Cedilnik]{Taskmaster2}
Bill Byrne, Karthik Krishnamoorthi, Saravanan Ganesh, Amit Dubey, Kyu-Young Kim, and Andy Cedilnik.
\newblock Taskmaster-2.
\newblock \url{https://github.com/google-research-datasets/Taskmaster/tree/master/TM-2-2020}, 2020.

\bibitem[Cao(2023)]{DiaggptAnLlmBasedChatbotWithAutomaticTopicManagementForTaskOrientedDialogue}
Lang Cao.
\newblock Diaggpt: An llm-based chatbot with automatic topic management for task-oriented dialogue.
\newblock \emph{CoRR}, abs/2308.08043, 2023.
\newblock \doi{10.48550/ARXIV.2308.08043}.
\newblock URL \url{https://doi.org/10.48550/arXiv.2308.08043}.

\bibitem[Chung et~al.(2023)Chung, Cahyawijaya, Wilie, Lovenia, and Fung]{InstructtodsLargeLanguageModelsForEndToEndTaskOrientedDialogueSystems}
Willy Chung, Samuel Cahyawijaya, Bryan Wilie, Holy Lovenia, and Pascale Fung.
\newblock Instructtods: Large language models for end-to-end task-oriented dialogue systems.
\newblock \emph{CoRR}, abs/2310.08885, 2023.
\newblock \doi{10.48550/ARXIV.2310.08885}.
\newblock URL \url{https://doi.org/10.48550/arXiv.2310.08885}.

\bibitem[Dalton et~al.(2020)Dalton, Xiong, Kumar, and Callan]{dalton2020cast}
Jeffrey Dalton, Chenyan Xiong, Vaibhav Kumar, and Jamie Callan.
\newblock Cast-19: A dataset for conversational information seeking.
\newblock In \emph{Proceedings of the 43rd International ACM SIGIR Conference on Research and Development in Information Retrieval}, pp.\  1985--1988, 2020.

\bibitem[Dettmers et~al.(2023)Dettmers, Pagnoni, Holtzman, and Zettlemoyer]{dettmers2023qlora}
Tim Dettmers, Artidoro Pagnoni, Ari Holtzman, and Luke Zettlemoyer.
\newblock Qlora: Efficient finetuning of quantized llms, 2023.

\bibitem[Feng et~al.(2021)Feng, Patel, Wan, and Joshi]{Multidoc2DialModelingDialoguesGroundedInMultipleDocuments}
Song Feng, Siva~Sankalp Patel, Hui Wan, and Sachindra Joshi.
\newblock Multidoc2dial: Modeling dialogues grounded in multiple documents.
\newblock In Marie{-}Francine Moens, Xuanjing Huang, Lucia Specia, and Scott~Wen{-}tau Yih (eds.), \emph{Proceedings of the 2021 Conference on Empirical Methods in Natural Language Processing, {EMNLP} 2021, Virtual Event / Punta Cana, Dominican Republic, 7-11 November, 2021}, pp.\  6162--6176. Association for Computational Linguistics, 2021.
\newblock \doi{10.18653/V1/2021.EMNLP-MAIN.498}.
\newblock URL \url{https://doi.org/10.18653/v1/2021.emnlp-main.498}.

\bibitem[Fu et~al.(2020)Fu, Xian, Zhu, Zhang, and de~Melo]{fu2020cookie}
Zuohui Fu, Yikun Xian, Yaxin Zhu, Yongfeng Zhang, and Gerard de~Melo.
\newblock Cookie: A dataset for conversational recommendation over knowledge graphs in e-commerce.
\newblock \emph{arXiv preprint arXiv:2008.09237}, 2020.

\bibitem[Gasic et~al.(2014)Gasic, Kim, Tsiakoulis, Breslin, Henderson, Szummer, Thomson, and Young]{IncrementalOnlineAdaptationOfPOMDPbased}
Milica Gasic, Dongho Kim, Pirros Tsiakoulis, Catherine Breslin, Matthew Henderson, Martin Szummer, Blaise Thomson, and Steve~J. Young.
\newblock Incremental on-line adaptation of pomdp-based dialogue managers to extended domains.
\newblock In Haizhou Li, Helen~M. Meng, Bin Ma, Engsiong Chng, and Lei Xie (eds.), \emph{{INTERSPEECH} 2014, 15th Annual Conference of the International Speech Communication Association, Singapore, September 14-18, 2014}, pp.\  140--144. {ISCA}, 2014.
\newblock \doi{10.21437/INTERSPEECH.2014-40}.
\newblock URL \url{https://doi.org/10.21437/Interspeech.2014-40}.

\bibitem[Hayati et~al.(2020)Hayati, Kang, Zhu, Shi, and Yu]{hayati2020inspired}
Shirley~Anugrah Hayati, Dongyeop Kang, Qingxiaoyang Zhu, Weiyan Shi, and Zhou Yu.
\newblock Inspired: Toward sociable recommendation dialog systems.
\newblock \emph{arXiv preprint arXiv:2009.14306}, 2020.

\bibitem[He \& McAuley(2016)He and McAuley]{amazon-reviews}
Ruining He and Julian McAuley.
\newblock Ups and downs: Modeling the visual evolution of fashion trends with one-class collaborative filtering.
\newblock In \emph{Proceedings of the 25th International Conference on World Wide Web}, WWW '16, pp.\  507–517, Republic and Canton of Geneva, CHE, 2016. International World Wide Web Conferences Steering Committee.
\newblock ISBN 9781450341431.
\newblock \doi{10.1145/2872427.2883037}.
\newblock URL \url{https://doi.org/10.1145/2872427.2883037}.

\bibitem[Hudecek \& Dusek(2023)Hudecek and Dusek]{AreLargeLanguageModelsAllYouNeedForTaskOrientedDialogue}
Vojtech Hudecek and Ondrej Dusek.
\newblock Are large language models all you need for task-oriented dialogue?
\newblock In David Schlangen, Svetlana Stoyanchev, Shafiq Joty, Ondrej Dusek, Casey Kennington, and Malihe Alikhani (eds.), \emph{Proceedings of the 24th Meeting of the Special Interest Group on Discourse and Dialogue, {SIGDIAL} 2023, Prague, Czechia, September 11 - 15, 2023}, pp.\  216--228. Association for Computational Linguistics, 2023.
\newblock URL \url{https://aclanthology.org/2023.sigdial-1.21}.

\bibitem[Jia et~al.(2022)Jia, Liu, Wang, Song, Xi, Li, Shen, Chen, Pang, and He]{jia2022convrec}
Meihuizi Jia, Ruixue Liu, Peiying Wang, Yang Song, Zexi Xi, Haobin Li, Xin Shen, Meng Chen, Jinhui Pang, and Xiaodong He.
\newblock E-convrec: a large-scale conversational recommendation dataset for e-commerce customer service.
\newblock In \emph{Proceedings of the Thirteenth Language Resources and Evaluation Conference}, pp.\  5787--5796, 2022.

\bibitem[Kang et~al.(2019)Kang, Balakrishnan, Shah, Crook, Boureau, and Weston]{kang2019recommendation}
Dongyeop Kang, Anusha Balakrishnan, Pararth Shah, Paul Crook, Y-Lan Boureau, and Jason Weston.
\newblock Recommendation as a communication game: Self-supervised bot-play for goal-oriented dialogue.
\newblock \emph{arXiv preprint arXiv:1909.03922}, 2019.

\bibitem[Kelley(1984)]{WizardOfOz}
J.~F. Kelley.
\newblock An iterative design methodology for user-friendly natural language office information applications.
\newblock \emph{{ACM} Trans. Inf. Syst.}, 2\penalty0 (1):\penalty0 26--41, 1984.
\newblock \doi{10.1145/357417.357420}.
\newblock URL \url{https://doi.org/10.1145/357417.357420}.

\bibitem[Leszczynski et~al.(2023)Leszczynski, Zhang, Ganti, Balog, Radlinski, Pereira, and Chaganty]{leszczynski2023talk}
Megan Leszczynski, Shu Zhang, Ravi Ganti, Krisztian Balog, Filip Radlinski, Fernando Pereira, and Arun~Tejasvi Chaganty.
\newblock Talk the walk: Synthetic data generation for conversational music recommendation.
\newblock \emph{arXiv preprint arXiv:2301.11489}, 2023.

\bibitem[Li et~al.(2018)Li, Ebrahimi~Kahou, Schulz, Michalski, Charlin, and Pal]{li2018towards}
Raymond Li, Samira Ebrahimi~Kahou, Hannes Schulz, Vincent Michalski, Laurent Charlin, and Chris Pal.
\newblock Towards deep conversational recommendations.
\newblock \emph{Advances in neural information processing systems}, 31, 2018.

\bibitem[Li et~al.(2022)Li, Chen, Li, Wang, Qian, and Yan]{ControllableDialogueSimulationWithInContextLearning}
Zekun Li, Wenhu Chen, Shiyang Li, Hong Wang, Jing Qian, and Xifeng Yan.
\newblock Controllable dialogue simulation with in-context learning.
\newblock In Yoav Goldberg, Zornitsa Kozareva, and Yue Zhang (eds.), \emph{Findings of the Association for Computational Linguistics: {EMNLP} 2022, Abu Dhabi, United Arab Emirates, December 7-11, 2022}, pp.\  4330--4347. Association for Computational Linguistics, 2022.
\newblock \doi{10.18653/V1/2022.FINDINGS-EMNLP.318}.
\newblock URL \url{https://doi.org/10.18653/v1/2022.findings-emnlp.318}.

\bibitem[Liao et~al.(2021)Liao, Long, Zhang, Huang, and Chua]{liao2021mmconv}
Lizi Liao, Le~Hong Long, Zheng Zhang, Minlie Huang, and Tat-Seng Chua.
\newblock Mmconv: an environment for multimodal conversational search across multiple domains.
\newblock In \emph{Proceedings of the 44th international ACM SIGIR conference on research and development in information retrieval}, pp.\  675--684, 2021.

\bibitem[Lin et~al.(2020)Lin, Auvray, Elkind, Biswas, Fazel{-}Zarandi, Belgamwar, Chandra, Zhao, Metallinou, Chung, Zhu, Adhikari, and Hakkani{-}T{\"{u}}r]{DialogSimulationWithRealisticVariationsForTrainingGoalOrientedConversationalSystems}
Chien{-}Wei Lin, Vincent Auvray, Daniel Elkind, Arijit Biswas, Maryam Fazel{-}Zarandi, Nehal Belgamwar, Shubhra Chandra, Matt Zhao, Angeliki Metallinou, Tagyoung Chung, Charlie~Shucheng Zhu, Suranjit Adhikari, and Dilek Hakkani{-}T{\"{u}}r.
\newblock Dialog simulation with realistic variations for training goal-oriented conversational systems.
\newblock \emph{CoRR}, abs/2011.08243, 2020.
\newblock URL \url{https://arxiv.org/abs/2011.08243}.

\bibitem[Mendon{\c{c}}a et~al.(2023)Mendon{\c{c}}a, Pereira, Menezes, Cabarr{\~{a}}o, Farinha, Moniz, Carvalho, Lavie, and Trancoso]{DialogueQualityAndEmotionAnnotationsForCustomerSupportConversations}
John Mendon{\c{c}}a, Patr{\'{\i}}cia Pereira, Miguel Menezes, Vera Cabarr{\~{a}}o, Ana~C. Farinha, Helena Moniz, Jo{\~{a}}o~Paulo Carvalho, Alon Lavie, and Isabel Trancoso.
\newblock Dialogue quality and emotion annotations for customer support conversations.
\newblock \emph{CoRR}, abs/2311.13910, 2023.
\newblock \doi{10.48550/ARXIV.2311.13910}.
\newblock URL \url{https://doi.org/10.48550/arXiv.2311.13910}.

\bibitem[Mohapatra et~al.(2021)Mohapatra, Pandey, Contractor, and Joshi]{SimulatedChatsForBuildingDialogSystemsLearningToGenerateConversationsFromInstructions}
Biswesh Mohapatra, Gaurav Pandey, Danish Contractor, and Sachindra Joshi.
\newblock Simulated chats for building dialog systems: Learning to generate conversations from instructions.
\newblock In Marie{-}Francine Moens, Xuanjing Huang, Lucia Specia, and Scott~Wen{-}tau Yih (eds.), \emph{Findings of the Association for Computational Linguistics: {EMNLP} 2021, Virtual Event / Punta Cana, Dominican Republic, 16-20 November, 2021}, pp.\  1190--1203. Association for Computational Linguistics, 2021.
\newblock \doi{10.18653/V1/2021.FINDINGS-EMNLP.103}.
\newblock URL \url{https://doi.org/10.18653/v1/2021.findings-emnlp.103}.

\bibitem[Qin et~al.(2023)Qin, Pan, Chen, Liao, Yu, Zhang, Che, and Li]{EndToEndTaskOrientedSurvey}
Libo Qin, Wenbo Pan, Qiguang Chen, Lizi Liao, Zhou Yu, Yue Zhang, Wanxiang Che, and Min Li.
\newblock End-to-end task-oriented dialogue: {A} survey of tasks, methods, and future directions.
\newblock In Houda Bouamor, Juan Pino, and Kalika Bali (eds.), \emph{Proceedings of the 2023 Conference on Empirical Methods in Natural Language Processing, {EMNLP} 2023, Singapore, December 6-10, 2023}, pp.\  5925--5941. Association for Computational Linguistics, 2023.
\newblock URL \url{https://aclanthology.org/2023.emnlp-main.363}.

\bibitem[Radford et~al.(2019)Radford, Wu, Child, Luan, Amodei, and Sutskever]{GPT2}
Alec Radford, Jeff Wu, Rewon Child, David Luan, Dario Amodei, and Ilya Sutskever.
\newblock Language models are unsupervised multitask learners.
\newblock 2019.
\newblock URL \url{https://api.semanticscholar.org/CorpusID:160025533}.

\bibitem[Rastogi et~al.(2020)Rastogi, Zang, Sunkara, Gupta, and Khaitan]{TowardsScalableMultiDomainConversationalAgentsTheSchemaGuidedDialogueDataset}
Abhinav Rastogi, Xiaoxue Zang, Srinivas Sunkara, Raghav Gupta, and Pranav Khaitan.
\newblock Towards scalable multi-domain conversational agents: The schema-guided dialogue dataset.
\newblock In \emph{The Thirty-Fourth {AAAI} Conference on Artificial Intelligence, {AAAI} 2020, The Thirty-Second Innovative Applications of Artificial Intelligence Conference, {IAAI} 2020, The Tenth {AAAI} Symposium on Educational Advances in Artificial Intelligence, {EAAI} 2020, New York, NY, USA, February 7-12, 2020}, pp.\  8689--8696. {AAAI} Press, 2020.
\newblock \doi{10.1609/AAAI.V34I05.6394}.
\newblock URL \url{https://doi.org/10.1609/aaai.v34i05.6394}.

\bibitem[Raux et~al.(2005)Raux, Langner, Bohus, Black, and Esk{\'{e}}nazi]{LetsGoPublic}
Antoine Raux, Brian Langner, Dan Bohus, Alan~W. Black, and Maxine Esk{\'{e}}nazi.
\newblock Let's go public! taking a spoken dialog system to the real world.
\newblock In \emph{{INTERSPEECH} 2005 - Eurospeech, 9th European Conference on Speech Communication and Technology, Lisbon, Portugal, September 4-8, 2005}, pp.\  885--888. {ISCA}, 2005.
\newblock \doi{10.21437/INTERSPEECH.2005-399}.
\newblock URL \url{https://doi.org/10.21437/Interspeech.2005-399}.

\bibitem[Sener \& Yao(2019)Sener and Yao]{sener2019zero}
Fadime Sener and Angela Yao.
\newblock Zero-shot anticipation for instructional activities.
\newblock In \emph{The IEEE International Conference on Computer Vision (ICCV)}, 2019.

\bibitem[Shah et~al.(2018)Shah, Hakkani{-}T{\"{u}}r, Liu, and T{\"{u}}r]{BootstrappingANeuralConversationalAgentWithDialogueSelfPlayCrowdsourcingAndOnLineReinforcementLearning}
Pararth Shah, Dilek Hakkani{-}T{\"{u}}r, Bing Liu, and G{\"{o}}khan T{\"{u}}r.
\newblock Bootstrapping a neural conversational agent with dialogue self-play, crowdsourcing and on-line reinforcement learning.
\newblock In Srinivas Bangalore, Jennifer Chu{-}Carroll, and Yunyao Li (eds.), \emph{Proceedings of the 2018 Conference of the North American Chapter of the Association for Computational Linguistics: Human Language Technologies, {NAACL-HLT} 2018, New Orleans, Louisiana, USA, June 1-6, 2018, Volume 3 (Industry Papers)}, pp.\  41--51. Association for Computational Linguistics, 2018.
\newblock \doi{10.18653/V1/N18-3006}.
\newblock URL \url{https://doi.org/10.18653/v1/n18-3006}.

\bibitem[Ulmer et~al.(2024)Ulmer, Mansimov, Lin, Sun, Gao, and Zhang]{BootstrappingLlmBasedTaskOrientedDialogueAgentsViaSelfTalk}
Dennis Ulmer, Elman Mansimov, Kaixiang Lin, Justin Sun, Xibin Gao, and Yi~Zhang.
\newblock Bootstrapping llm-based task-oriented dialogue agents via self-talk.
\newblock \emph{CoRR}, abs/2401.05033, 2024.
\newblock \doi{10.48550/ARXIV.2401.05033}.
\newblock URL \url{https://doi.org/10.48550/arXiv.2401.05033}.

\bibitem[Wen et~al.(2017)Wen, Vandyke, Mrksic, Gasic, Rojas{-}Barahona, Su, Ultes, and Young]{ANetworkBasedEndToEndTrainableTaskOrientedDialogueSystem}
Tsung{-}Hsien Wen, David Vandyke, Nikola Mrksic, Milica Gasic, Lina~Maria Rojas{-}Barahona, Pei{-}Hao Su, Stefan Ultes, and Steve~J. Young.
\newblock A network-based end-to-end trainable task-oriented dialogue system.
\newblock In Mirella Lapata, Phil Blunsom, and Alexander Koller (eds.), \emph{Proceedings of the 15th Conference of the European Chapter of the Association for Computational Linguistics, {EACL} 2017, Valencia, Spain, April 3-7, 2017, Volume 1: Long Papers}, pp.\  438--449. Association for Computational Linguistics, 2017.
\newblock \doi{10.18653/V1/E17-1042}.
\newblock URL \url{https://doi.org/10.18653/v1/e17-1042}.

\bibitem[Williams et~al.(2013)Williams, Raux, Ramachandran, and Black]{TheDialogStateTrackingChallenge}
Jason~D. Williams, Antoine Raux, Deepak Ramachandran, and Alan~W. Black.
\newblock The dialog state tracking challenge.
\newblock In \emph{Proceedings of the {SIGDIAL} 2013 Conference, The 14th Annual Meeting of the Special Interest Group on Discourse and Dialogue, 22-24 August 2013, SUPELEC, Metz, France}, pp.\  404--413. The Association for Computer Linguistics, 2013.
\newblock URL \url{https://aclanthology.org/W13-4065/}.

\bibitem[Xu et~al.(2020)Xu, Moon, Liu, Liu, Shah, and Yu]{xu2020user}
Hu~Xu, Seungwhan Moon, Honglei Liu, Bing Liu, Pararth Shah, and Philip~S Yu.
\newblock User memory reasoning for conversational recommendation.
\newblock \emph{arXiv preprint arXiv:2006.00184}, 2020.

\bibitem[Yang et~al.(2022)Yang, Wang, Yu, Kulkarni, Sanghai, Shu, Elsas, and Kanagal]{mave}
Li~Yang, Qifan Wang, Zac Yu, Anand Kulkarni, Sumit Sanghai, Bin Shu, Jon Elsas, and Bhargav Kanagal.
\newblock Mave: A product dataset for multi-source attribute value extraction.
\newblock In \emph{Proceedings of the Fifteenth ACM International Conference on Web Search and Data Mining}, WSDM '22, pp.\  1256–1265, New York, NY, USA, 2022. Association for Computing Machinery.
\newblock ISBN 9781450391320.
\newblock \doi{10.1145/3488560.3498377}.
\newblock URL \url{https://doi.org/10.1145/3488560.3498377}.

\bibitem[Yang et~al.(2021)Yang, Li, and Quan]{UbarTowardsFullyEndToEndTaskOrientedDialogSystemWithGpt2}
Yunyi Yang, Yunhao Li, and Xiaojun Quan.
\newblock {UBAR:} towards fully end-to-end task-oriented dialog system with {GPT-2}.
\newblock In \emph{Thirty-Fifth {AAAI} Conference on Artificial Intelligence, {AAAI} 2021, Thirty-Third Conference on Innovative Applications of Artificial Intelligence, {IAAI} 2021, The Eleventh Symposium on Educational Advances in Artificial Intelligence, {EAAI} 2021, Virtual Event, February 2-9, 2021}, pp.\  14230--14238. {AAAI} Press, 2021.
\newblock \doi{10.1609/AAAI.V35I16.17674}.
\newblock URL \url{https://doi.org/10.1609/aaai.v35i16.17674}.

\bibitem[Yao et~al.(2022)Yao, Chen, Yang, and Narasimhan]{yao2022webshop}
Shunyu Yao, Howard Chen, John Yang, and Karthik Narasimhan.
\newblock Webshop: Towards scalable real-world web interaction with grounded language agents.
\newblock \emph{Advances in Neural Information Processing Systems}, 35:\penalty0 20744--20757, 2022.

\bibitem[Young et~al.(2022)Young, Xing, Pandelea, Ni, and Cambria]{FusingTaskOrientedAndOpenDomainDialoguesInConversationalAgents}
Tom Young, Frank Xing, Vlad Pandelea, Jinjie Ni, and Erik Cambria.
\newblock Fusing task-oriented and open-domain dialogues in conversational agents.
\newblock In \emph{Thirty-Sixth {AAAI} Conference on Artificial Intelligence, {AAAI} 2022, Thirty-Fourth Conference on Innovative Applications of Artificial Intelligence, {IAAI} 2022, The Twelveth Symposium on Educational Advances in Artificial Intelligence, {EAAI} 2022 Virtual Event, February 22 - March 1, 2022}, pp.\  11622--11629. {AAAI} Press, 2022.
\newblock \doi{10.1609/AAAI.V36I10.21416}.
\newblock URL \url{https://doi.org/10.1609/aaai.v36i10.21416}.

\bibitem[Zamani et~al.(2023)Zamani, Trippas, Dalton, and Radlinski]{CISbook}
Hamed Zamani, Johanne~R. Trippas, Jeff Dalton, and Filip Radlinski.
\newblock Conversational information seeking.
\newblock \emph{Foundations and Trends® in Information Retrieval}, 17\penalty0 (3-4):\penalty0 244--456, 2023.
\newblock ISSN 1554-0669.
\newblock \doi{10.1561/1500000081}.
\newblock URL \url{http://dx.doi.org/10.1561/1500000081}.

\bibitem[Zhang et~al.(2023)Zhang, Peng, Li, Zhou, and Meng]{SGPTODBuildingTaskBotsEffortlesslyViaSchemaGuidedLlmPrompting}
Xiaoying Zhang, Baolin Peng, Kun Li, Jingyan Zhou, and Helen Meng.
\newblock {SGP-TOD:} building task bots effortlessly via schema-guided {LLM} prompting.
\newblock In Houda Bouamor, Juan Pino, and Kalika Bali (eds.), \emph{Findings of the Association for Computational Linguistics: {EMNLP} 2023, Singapore, December 6-10, 2023}, pp.\  13348--13369. Association for Computational Linguistics, 2023.
\newblock URL \url{https://aclanthology.org/2023.findings-emnlp.891}.

\bibitem[Zhang et~al.(2020)Zhang, Takanobu, Zhu, Huang, and Zhu]{RecentAdvancesAndChallengesInTaskOrientedDialogSystems}
Zheng Zhang, Ryuichi Takanobu, Qi~Zhu, MinLie Huang, and XiaoYan Zhu.
\newblock Recent advances and challenges in task-oriented dialog systems.
\newblock \emph{Science China Technological Sciences}, 63\penalty0 (10):\penalty0 2011--2027, Oct 2020.
\newblock ISSN 1869-1900.
\newblock \doi{10.1007/s11431-020-1692-3}.
\newblock URL \url{https://doi.org/10.1007/s11431-020-1692-3}.

\bibitem[Zhou et~al.(2020)Zhou, Zhou, Zhao, Wang, and Wen]{zhou2020towards}
Kun Zhou, Yuanhang Zhou, Wayne~Xin Zhao, Xiaoke Wang, and Ji-Rong Wen.
\newblock Towards topic-guided conversational recommender system.
\newblock \emph{arXiv preprint arXiv:2010.04125}, 2020.

\bibitem[Zhu et~al.(2018)Zhu, Lu, Zheng, Guo, Zhang, Wang, and Yu]{SELFBLEU}
Yaoming Zhu, Sidi Lu, Lei Zheng, Jiaxian Guo, Weinan Zhang, Jun Wang, and Yong Yu.
\newblock Texygen: A benchmarking platform for text generation models.
\newblock In \emph{The 41st International ACM SIGIR Conference on Research \& Development in Information Retrieval}, SIGIR '18, pp.\  1097–1100, New York, NY, USA, 2018. Association for Computing Machinery.
\newblock ISBN 9781450356572.
\newblock \doi{10.1145/3209978.3210080}.
\newblock URL \url{https://doi.org/10.1145/3209978.3210080}.

\end{thebibliography}
\bibliographystyle{colm2024_conference}

\newpage
\appendix
\section{Appendix}
In this section we provide some additional details about the features of our synthetic dataset compared to existing TOD and conversational recommendation datasets, detailed definitions of intents and slots we used for our experiments, confussion matrices for the intent classification task and two sample conversations from our best fine-tuned models.

\begin{table}[h]
\small
\centering
\begin{tabular}{ |l||c|c|c|c|c|c|c|  }
\toprule
\textbf{Dataset} & \textbf{ARQ} & \textbf{ODQ} & \textbf{CQ} & \textbf{RC} & \textbf{RS} & \textbf{CLQ} & \textbf{ST} \\
\midrule
ReDial \citep{li2018towards} & \checkmark &   &   &   &   & \checkmark & \checkmark \\
GoRecDial \citep{kang2019recommendation} & \checkmark &   &   &   &   & \checkmark &   \\
TG-ReDial \citep{zhou2020towards} & \checkmark & \checkmark &   &   &   &   &   \\
MGConvRex \citep{xu2020user} & \checkmark & \checkmark &   &   &   & \checkmark & \checkmark \\
INSPIRED \citep{hayati2020inspired} & \checkmark & \checkmark &   &   &   & \checkmark & \checkmark \\
MMConv \citep{liao2021mmconv} & \checkmark &   &   &   &   & \checkmark & \checkmark \\
COOKIE \citep{fu2020cookie} & \checkmark &   &   &   &   & \checkmark & \checkmark \\
E-ConvRec \citep{jia2022convrec} & \checkmark &   &   &   &   & \checkmark & \checkmark \\
MultiWOZ \citep{MultiWozALargeScaleMultiDomainWIzardOfOZDatasetForTaskOrientedDialogueModelling} & \checkmark & \checkmark &   &   &   & \checkmark & \checkmark \\
CAsT 19 \citep{dalton2020cast} & \checkmark &   & \checkmark &   &   & \checkmark & \checkmark \\
WebShop \citep{yao2022webshop} & \checkmark & \checkmark &   & \checkmark &   &   & \checkmark \\
SGD \citep{TowardsScalableMultiDomainConversationalAgentsTheSchemaGuidedDialogueDataset} & \checkmark &  &   & \checkmark &   & \checkmark & \checkmark \\
Synthetic Music Data \citep{leszczynski2023talk} & \checkmark &   &   &   &   & \checkmark & \checkmark \\
SynTOD (this work) & \checkmark & \checkmark & \checkmark & \checkmark & \checkmark & \checkmark & \checkmark \\
\bottomrule
\end{tabular}
\caption{Comparison of our dataset derived using the SynTOD framework with existing TOD and conversational recommendation datasets. Dataset features: ARQ (Attribute Related Question), ODQ (Open Domain Question), CQ (Comparison Questions), RC (Retrieval Component), RS (Result Summarization), CLQ (Clarifying Questions), ST (State Tracking).}
\label{tab:dataset-features}
\end{table}

\begin{figure}[h]
    \centering
    \includegraphics[width=\textwidth]{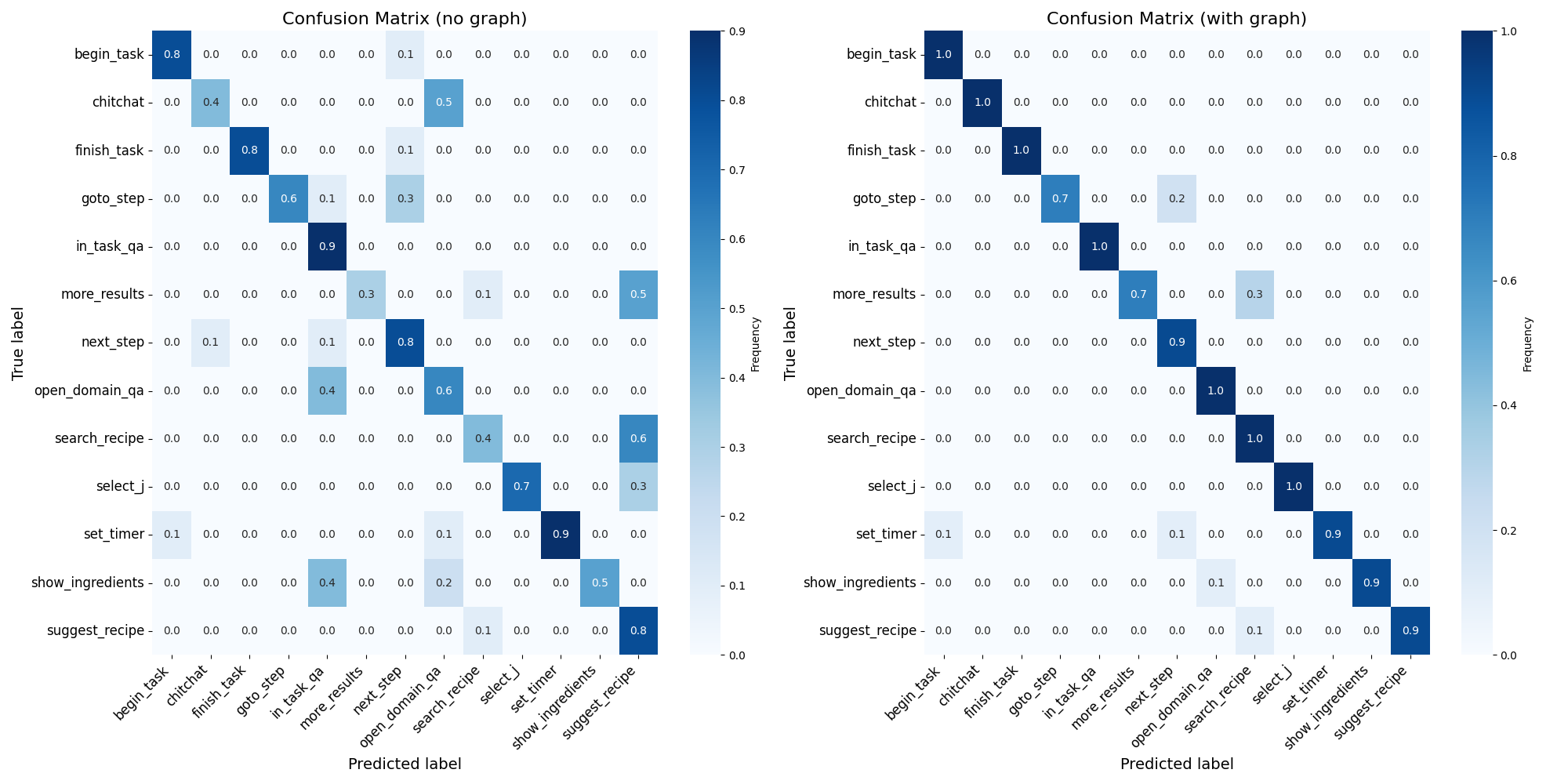}
    \caption{Intent classification confusion matrix for the cooking assistance domain without and using a state transition graph with the SynTOD framework. Using the graph-guided data generation approach, we observe significant improvements in intent classification across all intents.}
    \label{fig:recipe_domain_matrix}
\end{figure}

\begin{figure}[h]
    \centering
    \includegraphics[width=\textwidth]{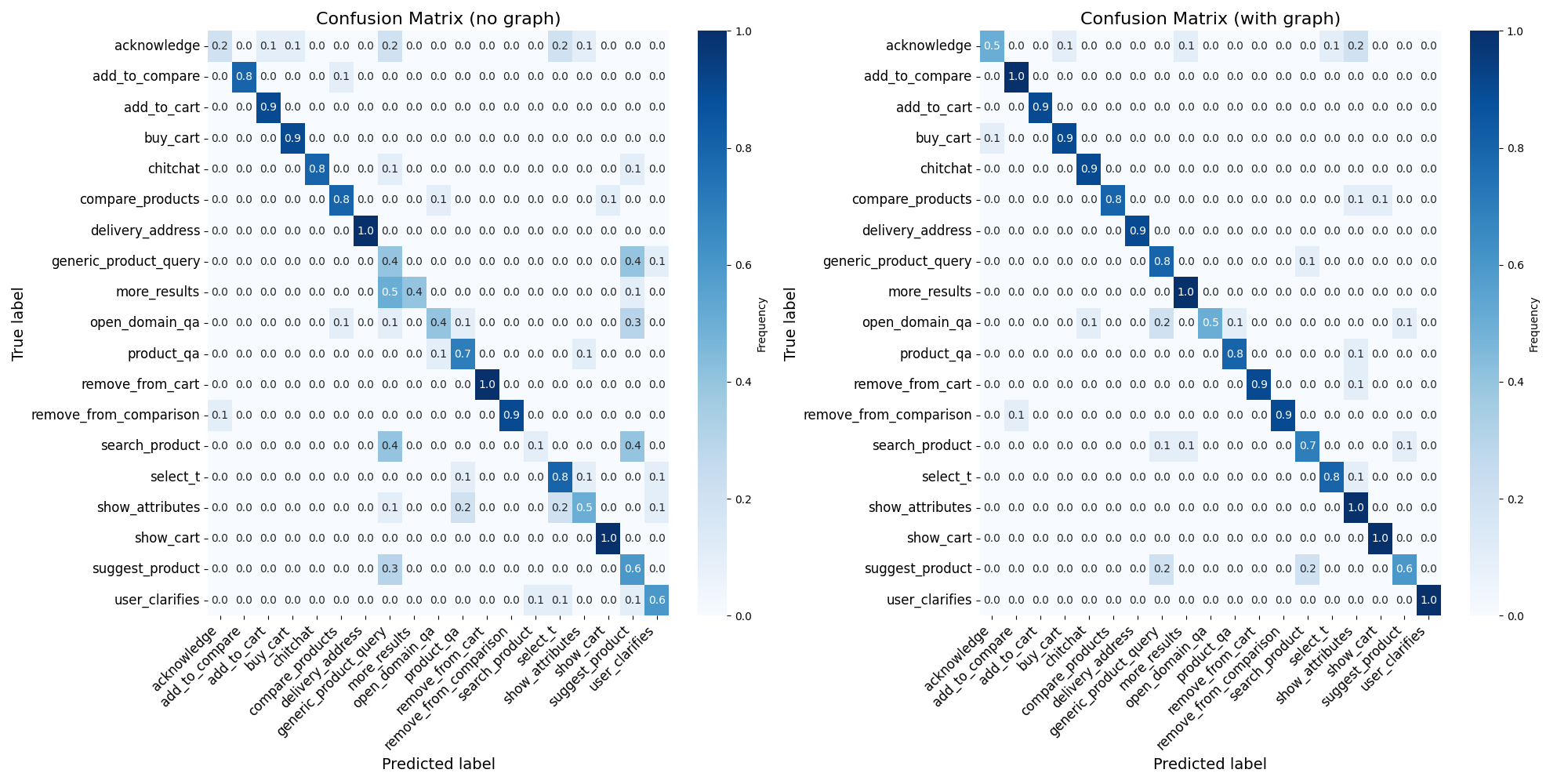}
    \caption{Intent classification confusion matrix for the e-commerce domain. Like in the cooking assistance domain, using the graph-guided data generation approach, we observe significant improvements in intent classification across all intents.}
    \label{fig:ecommerce_domain_matrix}
\end{figure}

\begin{table}[h!]
\small
\centering
\begin{tabular}{ll}
\toprule
\textbf{User Intent (slots)} & \textbf{Definition}  \\
\midrule
search\_recipe(query) & User searches for a specific recipe \\
suggest\_recipe(query) & User asks for recipe recommendation based on occasion, cuisine etc. \\
more\_results & Shows more recipe results \\
select(i) & Selects a recipe from a list \\
show\_ingredients & Asks to display the ingredients for a recipe \\
begin & Starts the recipe guidance \\
next\_step & Requests the next step in the recipe \\
goto\_step(i) & Jumps to a specific step in the recipe \\
repeat & Asks to repeat the current step or a previous step \\
finish\_task & Finishes the recipe guidance \\
open\_domain\_qa & Asks open-domain questions \\
in\_task\_qa & Asks questions related to recipe instructions \\
chitchat & Engages in general conversation not related to the task \\
set\_timer(duration) & Requests to set a timer for a specific duration \\
\bottomrule
\end{tabular}
\caption{Cooking Assistance User Intents and Slots we defined for our experiments.}
\label{tab:recipe-domain-intents}
\end{table}

\begin{table}[h!]
\small
\centering
\begin{tabular}{ll}
\toprule
\textbf{User Intent (slots)} & \textbf{Definition}  \\

\midrule
search\_product(query, attributes\_list) & Searching a specific product \\
suggest\_product(query, attributes\_list) & Asking about some product \\
generic\_product\_query & Very generic product query that needs clarification \\
user\_clarifies(query, attributes\_list) & Clarifying bot's question regarding a product. \\
more\_results & Asking bot to present more results \\
select(i) & Selecting a product from the display \\
show\_attributes(product\_id) & Asking to describe general features of a product \\
acknowledge & Acknowledging the reply from taskbot \\
open\_domain\_qa(question) & User asking an open domain question \\
product\_qa & Asking questions related to a product \\
chitchat & Engages in general conversation \\
show\_cart & Asking to show the cart \\
buy\_cart &  Asking to buy products present in the cart \\
add\_to\_cart(product\_id) & Asking to add a product to the cart \\
remove\_from\_cart(product\_id) & Asking to remove a product from the cart \\
compare\_products & Asking to compare products in compare list \\
delivery\_address(address) & Asking availability of a product in certain country \\
remove\_from\_comparison(product\_id) & Asking to remove products from compare list \\
add\_for\_comparison(product\_id) & Asking to add product in compare list \\

\bottomrule
\end{tabular}
\caption{Ecommerce Assistance User Intents we defined for our experiments.}
\label{tab:ecommerce-dataset-intents}
\end{table}

\begin{table}[htbp]
\fontsize{7.5}{1}\selectfont
\centering
\begin{tabular}{p{12cm}}
\toprule
\textbf{system}: Welcome to MarunaChef, your friendly recipe taskbot developed by CIIR at UMass Amherst! Whether you're a beginner cook or an experienced chef, I'm here to assist you in finding and preparing delicious recipes. From quick weekday meals to impressive dishes for special occasions, I've got you covered. Just let me know what ingredients you have or the dish you're craving, and I'll provide you with step-by-step instructions to create culinary delights. Let's get cooking! \textbf{intent:} start \\
\midrule
\textbf{user}: I'm having a party this weekend and I want to serve some amazing dips. Can you help me find some unique and easy dip recipes? \textbf{intent:} search\_recipe \\
\midrule
\textbf{system}: Here are some delicious dip recipes you might enjoy!\newline 1. **Taco Dip**: A spicy and savory party appetizer that's a crowd-pleaser, with a high rating of 4.7. It's loaded with melty cheese, spicy poblanos, and savory beef. Ready in just 25 minutes!\newline 2. **Buffalo Chicken Dip**: This creamy dip is a fan favorite, boasting a stellar rating of 4.9. It's packed with shredded chicken, hot sauce, and plenty of melty cheese. Also ready in 25 minutes!\newline3. **Chocolate Brownie Dip**: For the sweet tooths out there, this decadent dip is perfect for sharing or indulging solo. It's a quick 5-minute recipe with a rating of 3.5.\newline Which one tickles your taste buds? Let me know if you'd like to see the full recipe for any of these, or if you'd like more options. \textbf{intent:} show\_results \\
\midrule
\textbf{user}: I'd like to go for the recipe with the highest rating please. \textbf{intent:} select\_i \\
\midrule
\textbf{system}: Great choice! Buffalo Chicken Dip is a classic crowd-pleaser. Would you like me to show you the list of required ingredients? And when you're ready, just let me know, and we'll dive right into the first step. \textbf{intent:} option\_selected \\
\midrule
\textbf{user}: What is the weather like in London right now? \textbf{intent:} acknowledge\_task \\
\midrule
\textbf{system}: I'm sorry for any confusion, but as an AI developed by CIIR at UMass Amherst, I don't have real-time data capabilities and cannot provide current weather updates. I recommend checking a trusted weather service for the most accurate information. \textbf{intent:} started\_task \\
\midrule
\textbf{user}: This recipe seems interesting, let's not waste any more time and get started. \textbf{intent:} acknowledge\_step \\
\midrule
\textbf{system}: Alright, let's dive right in! The first step is to preheat your oven to 350°F (180°C). While it's warming up, we can proceed with the next steps. \textbf{intent:} show\_step \\
\midrule
\textbf{user}: The oven is preheated. Now what?  \textbf{intent:} done\_step \\
\midrule
\textbf{system}: In step 2 of the recipe, you'll be mixing together the shredded chicken, cream cheese, and buffalo sauce. This combination creates a creamy and spicy filling for the dip. It's important to evenly distribute the ingredients for the perfect flavor. When you're ready, let me know and we can move on to the next step!  \textbf{intent:} show\_step \\
\midrule
\textbf{user}: I'm done mixing the chicken, cream cheese, and buffalo sauce. \textbf{intent:} done\_step \\
\midrule
\textbf{system}: Step 3 of the Buffalo Chicken Dip recipe is all about getting that mixture into a cast iron skillet and making it smooth with a spoon. This step helps to distribute the flavors evenly and ensures a creamy texture throughout. So, when you're ready to continue, let me know! And here's a fun tip: Using a cast iron skillet allows for even heating, giving your dip a lovely golden crust. Enjoy cooking!  \textbf{intent:} show\_step \\
\midrule
\textbf{user}: What exactly does the fifth step of this recipe involve? \textbf{intent:} goto\_step\_5 \\
\midrule
\textbf{system}:  Step 5 of this mouthwatering recipe is where the magic happens! After it's baked to perfection, you get to top your Buffalo Chicken Dip with fresh scallions for an extra burst of flavor. Don't forget to grab your favorite chips for dipping! Are you ready to continue, or would you like a tip or fun fact? \textbf{intent:} no\_more\_steps \\
\midrule
\textbf{user}:  I've followed the last step and the dip is complete. \textbf{intent:} finish\_task \\
\midrule
\textbf{system}: Congratulations on completing the mouthwatering Buffalo Chicken Dip! You did an incredible job following the recipe! Is there anything else I can assist you with? Feel free to ask any further questions or explore more delicious recipes. Happy cooking! \textbf{intent:} start \\
\midrule
\textbf{user}:  That's all I needed, I'll let you know how it turns out! \textbf{intent:} task\_complete \\
\midrule
\textbf{system}: Great job! I hope you enjoyed making this recipe. Enjoy your meal and don't hesitate to reach out if you'd like to try another recipe. Bon Appétit! \textbf{intent:} end \\
\bottomrule
\end{tabular}
\caption{Cooking Assistance Sample Conversation}
\label{tab:recipe-dataset-sample-conversation}
\end{table}

\begin{table}[htbp]
\fontsize{7}{1}\selectfont
\centering
\begin{tabular}{p{\textwidth}}
\toprule
\textbf{system}: Hello! Welcome to MarunaShopper, your personal taskbot designed by CIIR at UMass Amherst to assist you in finding the perfect items in our product catalog. How can I help you shop today? \textbf{intent:} start \\
\midrule
\textbf{user}: I'm looking for something to protect my iPod with a bit of style. Can you help me find a suitable case? \textbf{intent:} generic\_product\_query \\
\midrule
\textbf{system}: Absolutely, I'd be happy to help you find a stylish case for your iPod! To narrow down the options, could you tell me which iPod model you have? Also, do you have any preferences when it comes to material, color, or any specific design elements like patterns or prints? Would you like any additional features such as a stand, pockets for cards, or extra shock absorption? \textbf{intent:} clarifying\_questions \\
\midrule
\textbf{user}: Oh, I have the 4th Generation iPod touch and I love a bit of uniqueness in my accessories. I'm really into animal prints and my favorite color combo is black and white. Something with a bit of bling wouldn't hurt either, as long as it's not too over the top. And I need good scratch protection, of course. Could you find me something like that? \textbf{intent:} user\_clarifies \textbf{query:} iPod touch 4th Generation case black white animal print bling \textbf{attributes\_list:} [\{"Color": "Black and White"\}, \{"Compatibility":"iPod touch 4th Gen"\}] \\
\midrule
\textbf{system}: Take a look at our stylish options for your iPod touch 4th Generation case needs! 1. Step up your game with the *eForCity Snap-On Rubber Coated Case for iPod touch 4G (Zebra/Bling Bow and Heart)*. This eye-catching case not only provides durable... It's a perfect fit for your device and allows easy ... 2. For those who love patterns, the *Green Zebra TPU Rubber Skin Case Cover for iPod Touch 4th Generation 4G 4* might be... Feel free to add any of these cases to your cart... iPod touch a new look while ensuring it stays protected. Which one catches your eye? \textbf{intent:} show\_results \\
\midrule
\textbf{user}: The Snap-On case with the rubber coating and that unique design is my choice. \textbf{intent:} select\_i \\
\midrule
\textbf{system}: Absolutely, the Zebra Bling Case for the iPod touch 4G sounds like a stylish choice! You can take a closer look at its features, add it to your cart for purchase, or if you're still deciding, you can compare it with other cases. What would you like to do next? \textbf{intent:} option\_selected \\
\midrule
\textbf{user}: Include the eForCity Snap-On Case with Zebra and Bling Bow design in the comparison list to see how it stacks up against the other options. \textbf{intent:} add\_for\_compare \textbf{title:} eForCity Snap-On Rubber Coated Case for iPod touch 4G \textbf{product\_id:} 3 \\
\midrule
\textbf{system}: Certainly! I've added the eForCity Snap-On Rubber Coated Case for the iPod touch 4G with the chic Zebra and Bling Bow design to your comparison list. You can now easily compare its style, durability, and user rating against your other choices. If you need detailed comparisons on specific attributes like compatibility, protection level, or how it might complement your style, feel free to ask! \textbf{intent:} system\_response\_add\_for\_compare \\
\midrule
\textbf{user}: Can you suggest a protective case with a unique design for my Apple iPod Touch 4th Generation that also includes a screen protector and has high user ratings? \textbf{intent:} suggest\_product \textbf{query:} iPod Touch 4th Gen case screen protector high rating \textbf{attributes\_list:} [\{Compatibility	:	Apple iPod\}, \{Feature	:	Screen Protector\}, \{Rating	:	5.0\} ] \\
\midrule
\textbf{system}: Discover the perfect companions... Feel free to add any of these to your cart, compare them side by side, or proceed to purchase to give your iPod Touch 4th Gen the protection it deserves. \textbf{intent:} show\_results \\
\midrule
\textbf{user}: I'd like to choose the protector that comes with a cleaning cloth and has precise cutouts for the buttons and ports. \textbf{intent:} select\_i \\
\midrule
\textbf{system}: Great choice! The Purple Butterfly Tree Case for the iPod Touch 4 offers both style and protection. You can view its detailed attributes, add it to your cart for later, proceed to purchase it now, or add it to your comparison list. What would you like to do next? \textbf{intent:} option\_selected \\
\midrule
\textbf{user}: Can you provide details on the material composition of the hard plastic used in the Snap on Hard Cover Protector Case, and whether the rubberized coating extends to all surfaces of the case? \textbf{intent:} product\_qa \textbf{product\_id:} 2 \\
\midrule
\textbf{system}: Certainly! The *Purple Butterfly Tree Polka Rubberized Snap on Hard Cover Protector Faceplate Case* is crafted with high-quality..., protecting your Apple Ipod Touch 4th Generation from scratches, dents, and chips. The hard plastic is reinforced... Would you be interested in adding this protective and stylish case to your cart, or is there anything else I can assist you with today? \textbf{intent:} product\_qa\_system\_response \\
\midrule
\textbf{user}: Include the Purple Butterfly Tree Case in the comparison list to evaluate it alongside the other options. \textbf{intent:} add\_for\_compare \textbf{title:} Purple Butterfly Tree Polka Rubberized Snap on Hard Cover Protector Faceplate Case
\textbf{product\_id:} 2 \\
\midrule
\textbf{system}: Absolutely, the Purple Butterfly Tree Polka Rubberized Snap-on Hard Cover Protector Faceplate Case for the Apple iPod Touch 4th Generation will be added to your comparison list... If you need further assistance or have more items to compare, please let me know! \textbf{intent:} system\_response\_add\_for\_compare \\
\midrule
\textbf{user}: Can you provide a detailed comparison between these iPod touch cases, focusing on their compatibility, style, protection level, and user ratings? \textbf{intent:} compare\_products \\
\midrule
\textbf{system}: Certainly! Let's compare the two iPod Touch 4th Generation cases you're interested in: *Purple Butterfly Tree Polka Rubberized Snap on Hard Cover Protector Faceplate Case for Apple Ipod Touch 4 4th Gen.* - This case offers... - It is designed to protect your device from scratches, dents, and chips while... \textbf{intent:} show\_comparison \textbf{product\_ids:} [1,2] \\
\midrule
\textbf{user}: Please remove the item with the Purple Butterfly Tree design from our comparison list. \textbf{intent:} select\_i\_remove\_from\_compare \textbf{title:} Purple Butterfly Tree Polka Rubberized Snap on Hard Cover Protector Faceplate Case
\textbf{product\_id:} 1 \\
\midrule
\textbf{system}: Absolutely, the Purple Butterfly Tree Polka Rubberized Snap ... has been removed from your comparison list. If there's anything else... \textbf{intent:} system\_response\_remove\_from\_compare \\
\midrule
\textbf{user}: Could you find me the South Korea flag patch with a gold border that can be ironed on? \textbf{intent:} search\_product \\
\midrule
... \\
\midrule
\textbf{user}: Please add the South Korea Flag patch to my shopping cart. \textbf{intent:} add\_to\_cart \\
\midrule
\textbf{system}: ... \textbf{intent:} system\_response\_added\_to\_cart \\
\midrule
\textbf{user}: Could you please show me the items currently in my cart? \textbf{intent:} show\_cart \\
\midrule
\textbf{system}: Here are the items in your shopping cart so far: 1. *SOUTH KOREA FLAG w/GOLD BORDER/Iron On Patch Applique/South Korean National Flag* Please let me know if you would like to proceed to checkout or continue shopping! \textbf{intent:} shown\_cart \\
\midrule
\textbf{user}: I'm ready to make the purchase. \textbf{intent:} buy\_cart \\
\bottomrule
\end{tabular}
\caption{E-commerce Assistance Sample Conversation.}
\label{tab:ecommerce-dataset-sample-conversation}
\end{table}

\end{document}